\documentclass{article}
\usepackage[final]{colm2026_conference}

\usepackage[T1]{fontenc}

\usepackage[ruled,linesnumbered,vlined]{algorithm2e}
\usepackage{amsmath,amssymb}
\usepackage{booktabs}
\usepackage{graphicx}
\usepackage{microtype}
\usepackage{hyperref}
\usepackage{url}
\usepackage{float}
\usepackage{tikz}
\usetikzlibrary{arrows.meta,calc,decorations.pathreplacing,positioning}
\usepackage{enumitem}
\usepackage{listings}
\usepackage{lineno}
\usepackage[most]{tcolorbox}
\usepackage{wasysym}
\usepackage{pifont}
\usepackage{caption}
\newcommand{\cmark}{\ding{51}}
\newcommand{\xmark}{\ding{55}}

\definecolor{pathA}{HTML}{2060A0}   % Wikidata path (crossdomain)
\definecolor{pathB}{HTML}{A02020}   % entity-universe path

\SetAlFnt{\footnotesize}
\SetAlCapFnt{\footnotesize}
\SetAlCapNameFnt{\footnotesize}

\title{\textsc{DrBencher}: Can Your Agent Identify the Entity, Retrieve Its Properties \emph{and} Do the Math?}
\author{Young-Suk Lee, Ram\'{o}n Fernandez Astudillo \& Radu Florian \\
IBM Research \\
\texttt{\{ysuklee,raduf\}@us.ibm.com}, \texttt{ramon.astudillo@ibm.com}}

\begin{document}

\ifcolmsubmission
\linenumbers
\fi

\maketitle

\begin{abstract}
Deep research agents increasingly interleave web browsing with
multi-step computation, yet existing benchmarks evaluate these
capabilities in isolation, creating a blind spot in assessing
real-world performance.  We introduce \textsc{DrBencher}, a synthetic
benchmark generator for questions that require both browsing and
computation, synthesized \emph{answer-first} from knowledge-graph
chains with no seed passages or provenance text. It enforces four
criteria: \emph{verifiability} (gold
answers are computed by executing parameterized code over
knowledge-graph values), \emph{complexity} (multi-hop entity
identification, property retrieval, and domain-specific computation),
\emph{difficulty} (a two-stage verification cascade filters out
questions solvable by the generating model), and \emph{diversity} (a
greedy max-min embedding filter maximizes coverage).  These criteria
are realized via a unified pipeline spanning five
domains: biochemistry, financial, geophysical, security, and history.
Human evaluation shows 76\% validity with
35\% of errors due to outdated knowledge-graph entries, highlighting
an inherent limitation of systems that reason over evolving data.
Automatic evaluation shows that the strongest frontier model achieves
only 20\% answer accuracy.  Compared to manually constructed
benchmarks (BrowseComp+, MATH-500, GPQA), \textsc{DrBencher} achieves
the highest semantic diversity. We release \textsc{DrBencher} pipeline and the human-verified benchmark data at \url{https://github.com/IBM/DrBencher}.
\end{abstract}

\section{Introduction}
\label{sec:introduction}

Large language models (LLMs) have rapidly advanced from single-turn question answering to multi-step deep research agents that orchestrate web search, code execution, and iterative reasoning to answer complex questions~\citep{yao2023react, mialon2024gaia}. Evaluating such agents demands benchmarks whose questions simultaneously require browsing and computation.
Curated multi-hop datasets such as HotpotQA~\citep{yang2018hotpotqa}, MuSiQue~\citep{trivedi2022musique}, and 2WikiMultihopQA~\citep{ho2020constructing} were constructed for a prior generation of models and are increasingly solvable by current systems. The recent BrowseComp~\citep{wei2025browsecomp} and FRAMES~\citep{krishna2025frames} benchmarks target agentic browsing and multi-hop factual retrieval, yet their questions focus on a single skill, i.e.\ browsing. Static benchmarks such as MMLU~\citep{hendrycks2021mmlu} and MATH~\citep{hendrycks2021math} risk contamination through training-set memorization \cite{yang2023rethinking,zhao2025mmlucf}. Answers to fixed-set benchmarks leak onto the public web; frontier models have been observed to independently identify which benchmark they are running and locate the answer key~\citep{coleman2026evalaware}.

%Meanwhile, SimpleQA~\citep{wei2024simpleqa} and FreshQA~\citep{vu2024freshqa} focus on factual accuracy but do not test the compositional, multi-step reasoning chains that distinguish deep research from simple lookup.

% ── Pipeline Figure ──
\begin{figure*}[!t]
\centering
\resizebox{\textwidth}{!}{%
\begin{tikzpicture}[
    >=Stealth,
    node distance=0.3cm,
    % --- box styles ---
    phase/.style={draw, rounded corners=4pt, text width=14.5cm,
                  align=left, font=\small, inner sep=6pt},
    gen/.style={phase, fill=blue!6, draw=blue!45},
    llm/.style={phase, fill=orange!8, draw=orange!45},
    ver/.style={phase, fill=red!6, draw=red!35},
    div/.style={phase, fill=green!6, draw=green!40},
    io/.style={draw, rounded corners=3pt, fill=gray!10, draw=gray!45,
               minimum height=0.7cm, align=center, font=\small},
    eg/.style={font=\small\itshape, text=gray!70!black},
    crit/.style={font=\footnotesize\bfseries, rounded corners=2pt,
                 minimum height=0.5cm, inner sep=3pt},
    discard/.style={font=\small, text=red!65!black},
    arrow/.style={->, thick, draw=gray!65},
    darrow/.style={->, thick, draw=red!45, dashed},
]

% ── Input ──
\node[io, text width=14.5cm] (input) {Domain $D \in$ \{\textsc{bio}, \textsc{fin}, \textsc{geo}, \textsc{hist}, \textsc{sec}\},\; Topic $T \in$ \{mountains, tech, enzymes, CVEs, wars, \ldots\}};

% ── Phase 0 ──
\node[gen, below=0.55cm of input] (p0) {%
  \textbf{Phase 0 --- Seed Entity Discovery \& Gold Answer Computation}\\[1pt]
  {\footnotesize
    \begin{tabular}[t]{@{}r@{\;}l@{}}
    $\bullet$ & Query Wikidata for typed entities
      \quad{\small\itshape\color{gray!70!black}$\to$ Mount Fuji}\\
    $\bullet$ & Fetch multi-hop KG chains + domain data
      \quad{\small\itshape\color{gray!70!black}$\to$ Fuji $\to$ located in $\to$ Shizuoka $\to$ \ldots}\\
    $\bullet$ & Select reasoning template
      \quad{\small\itshape\color{gray!70!black}$\to$ atmospheric\_pressure}\\
    $\bullet$ & Execute code $\to$ gold answer $a^*$
      \quad{\small\itshape\color{gray!70!black}$\to$ ``63.4 kPa''}\\
    \end{tabular}}};

% ── Phase 1 ──
\node[llm, below=of p0] (p1) {%
  \textbf{Phase 1 --- Entity Clue Derivation from KG Chains}\\[1pt]
  {\footnotesize
    \begin{tabular}[t]{@{}r@{\;}l@{}}
    $\bullet$ & Derive natural-language description per chain about the seed entity\\
    $\bullet$ & Require clues spanning $\ge$3 distinct chains\\
    \end{tabular}}\\[2pt]
  {\small\itshape\color{gray!70!black}e.g.\; Clue 1: ``\textnormal{Located in a prefecture whose capital hosted the 1957 National Sports Festival}''\;\;
    Clue 2: ``\textnormal{Highest peak on an island where the Jomon period began}''}};

% ── Phase 1.5 ──
\node[llm, below=of p1] (p15) {%
  \textbf{Phase 1.5 --- Fact Verification against Wikipedia}\\[1pt]
  {\footnotesize
    \begin{tabular}[t]{@{}r@{\;}l@{}}
    $\bullet$ & For each derived fact, verify claim is supported by the corresponding Wikipedia article\\
    $\bullet$ & Reject ungrounded facts;\; require $\ge$3 survivors\\
    \end{tabular}}\\[2pt]
  {\small\itshape\color{gray!70!black}e.g.\; ``\textnormal{hosted the 1957 National Sports Festival}'' $\checkmark$ confirmed in Wikipedia on Shizuoka;\;\;
    ``\textnormal{famous for cherry blossoms}'' $\times$ rejected}};

% ── Phase 2 ──
\node[llm, below=of p15] (p2) {%
  \textbf{Phase 2 --- Question Composition}\\[1pt]
  {\footnotesize
    Compose a natural-language question from derived entity clues and the reasoning template}\\[2pt]
  {\small\itshape\color{gray!70!black}e.g.\; ``\textnormal{It is a mountain located in a prefecture whose capital hosted the 1957 National Sports Festival. It is the highest peak on an island where the Jomon period began. What is the atmospheric pressure at its summit in kPa?}''}};

% ── Phase 3: Programmatic QA Validation ──
\node[gen, below=of p2] (p3) {%
  \textbf{Phase 3 --- Programmatic QA Validation}\\[1pt]
  {\footnotesize
    \begin{tabular}[t]{@{}r@{\;}l@{}}
    $\bullet$ & Re-compute the answer (must match $a^*$)
      \quad{\small\itshape\color{gray!70!black}$\to$ \textnormal{63.4 kPa} $=$ $a^*$ $\checkmark$}\\
    $\bullet$ & Gold entity/value leak check in the question
      \quad{\small\itshape\color{gray!70!black}$\to$ ``\textnormal{Mount Fuji}'' not in question $\checkmark$}\\
    $\bullet$ & Question ambiguity check $\to$ unique answer
      \quad{\small\itshape\color{gray!70!black}$\to$ clues uniquely identify \textnormal{Mount Fuji} $\checkmark$}\\
    \end{tabular}}};

% ── Verification (V1/V2) ──
\node[ver, below=of p3] (vbox) {%
  \textbf{Two-Stage Difficulty Verification} --- ensure each QA pair is sufficiently challenging\\[3pt]
  {\footnotesize\textbf{V1}\; Closed-book:\;
    $k_1$ samples, no tools;\;
    discard if $\mathrm{acc} \ge \tau_1$\;
    {\small\itshape\color{gray!70!black}(too easy from parametric knowledge)}}\\[1pt]
  {\footnotesize\textbf{V2}\; Tool-augmented:\;
    $k_2$ samples with browser + Python tools;\;
    discard if $\mathrm{acc} \ge \tau_2$\;
    {\small\itshape\color{gray!70!black}(solvable with tools)}}};

% ── Diversity ──
\node[div, below=of vbox] (divf) {%
  \textbf{Diversity Filter}\\[1pt]
  {\footnotesize Graph-based max-min embedding filter
    selects maximally spread subset from survivors}};

% ── QA Validation by Claude Opus 4.6 ──
%\node[gen, below=of divf, inner sep=4pt] (cval) {%
%  \textbf{QA Validation by Claude Opus 4.6}};

% ── Output ──
\node[io, text width=14.5cm, below=0.55cm of divf] (output) {Benchmark $\mathcal{Q}^*$};

% ── Main flow arrows ──
\foreach \a/\b in {input/p0, p0/p1, p1/p15, p15/p2, p2/p3, p3/vbox, vbox/divf, divf/output}
  \draw[arrow] (\a) -- (\b);

% ── Discard arrows (right side) ──
\node[discard, right=0.6cm of p3.east, anchor=west] (d3) {$\times$\; discard};
\draw[darrow] (p3.east) -- (d3.west);

\node[discard, right=0.6cm of vbox.east, anchor=west] (dv) {$\times$\; discard};
\draw[darrow] (vbox.east) -- (dv.west);

\node[discard, right=0.6cm of divf.east, anchor=west] (dd) {$\times$\; discard};
\draw[darrow] (divf.east) -- (dd.west);

% ── Left-side criterion labels ──
% Verifiability bracket spanning Phases 0–2
\node[crit, fill=blue!10, anchor=east] at
  ([xshift=-0.9cm]$(p0.west)!0.5!(p2.west)$)
  {\textcolor{blue!65!black}{Verifiability}};
\draw[blue!40, thick, decorate, decoration={brace, amplitude=5pt, mirror}]
  ([xshift=-0.5cm]p0.north west) -- ([xshift=-0.5cm]p2.south west);

% Complexity label for Phase 2
\node[crit, fill=orange!12, left=0.9cm of p2.west, anchor=east]
  {\textcolor{orange!65!black}{Complexity}};

% Correctness label for Phase 3
\node[crit, fill=blue!10, left=0.9cm of p3.west, anchor=east]
  {\textcolor{blue!65!black}{Correctness}};

% Difficulty label for Verification box
\node[crit, fill=red!10, left=0.9cm of vbox.west, anchor=east]
  {\textcolor{red!55!black}{Difficulty}};

% Diversity label
\node[crit, fill=green!10, left=0.9cm of divf.west, anchor=east]
  {\textcolor{green!50!black}{Diversity}};

\end{tikzpicture}%
}% end resizebox
\caption{\small The unified \textsc{DrBencher} pipeline, shared across all five domains
(\textsc{biochemistry}, \textsc{financial}, \textsc{geophysical}, \textsc{history}, \textsc{security}).
A running example using \emph{Mount Fuji} (atmospheric pressure template) illustrates each stage.}
\label{fig:pipeline}
\vspace{-2pt}
\end{figure*}

There is a growing need for benchmarks that test multi-skill reasoning
and that can be instantiated fresh on demand, eliminating static answer keys
susceptible to leakage or reverse-engineering.  In this paper, we
propose \textsc{DrBencher} (Deep Research Benchmarker), a novel synthetic benchmark generator for
questions that require both browsing and computation.
Figure~\ref{fig:pipeline} illustrates the overall pipeline. We
identify four optimization axes and formalize them as criteria for
synthetic benchmark generation.  \textbf{Verifiability} concerns the
\emph{answer}: every gold answer is deterministically checkable by
executing parameterized code against KG-sourced
values~\citep{vrandecic2014wikidata}, eliminating subjective judgment.
\textbf{Complexity} concerns the \emph{question}: each question
demands entity identification from multi-hop KG clues, retrieval of
quantitative properties, and computation over those values---no stage
can be skipped. We quantify this with a \emph{Compositional Complexity
Index} (CCI $= E + P$), where $E$ counts entities and $P$ counts
property lookups. CCI is model-independent and correlates with
extrinsic difficulty (Table~\ref{tab:cci}).  \textbf{Difficulty:} a
two-stage self-referential filter discards questions solvable by the
generating model, either closed-book or with tool-augmented
agents~\citep{li2025autobencher}.  \textbf{Diversity:} a graph-based
max-min embedding filter~\citep{friedman2023vendi} maximizes coverage
of entities, templates, and linguistic expression.  We evaluate
\textsc{DrBencher} along three axes: \emph{validity}---human
annotators rate 76\% of question and answer pairs as valid (84\%
excluding stale data), with 35\% of all errors traced to outdated
knowledge-graph entries; \emph{difficulty}---the strongest frontier
model identifies entities 86\% of the time yet achieves only 20\%
accuracy, with property retrieval as the primary bottleneck; and
\emph{diversity}---compared to manually constructed benchmarks
(BrowseComp+, MATH-500, GPQA), \textsc{DrBencher} attains the highest
semantic diversity. 

We make the following new contributions:

\begin{itemize}[leftmargin=*]

\item \textbf{Multi-skill compositional questions.} 
%To our knowledge, this is the first benchmark generator that \emph{couples} multi-hop entity identification, property retrieval, and quantitative reasoning from knowledge-graph chains. -- LC-QuAD, KQA Pro  involve retrieving properties from KG chains, we might get complaints from reviewers for too broad of a claim (Hans)
To our knowledge, this is the first benchmark generator that combines
indirect identification of withheld entities from KG-chain clues with
domain-specific quantitative computation over their
properties. Existing multi-hop datasets~\citep{yang2018hotpotqa,
  trivedi2022musique, ho2020constructing, press2023bamboogle} test
retrieval composition but not mathematical or scientific
computation. Conversely, mathematical
benchmarks~\citep{hendrycks2021math, chen2021codex} test computation
but not information retrieval.

\item \textbf{No seed passages or provenance text.}
Unlike multi-hop datasets and synthetic-QA pipelines that derive
questions \emph{from} pre-existing source passages or seed documents,
\textsc{DrBencher} uses no seed passages or provenance text:
questions are synthesized answer-first from SPARQL-selected entities
and their KG chains, with Wikipedia used only to verify the grounding
of clue facts rather than as the source they are drawn from. Because each instance is generated on demand from live KG data with
  programmatically verified reference answers, the resulting benchmark
  is contamination-resistant by design, eliminating the static answer
  key that enables leakage and reverse-engineering in fixed-set
  benchmarks~\citep{coleman2026evalaware}.

\item \textbf{Benchmark generation as multi-criterion enforcement.} We
  frame synthetic benchmark construction around four jointly enforced
  criteria. Prior work on synthetic benchmarks---notably
  AutoBencher~\citep{li2025autobencher}---addresses topic coverage and
  difficulty but neither enforces programmatic verifiability nor
  controls multi-skill compositional complexity measured by
  model-independent metric \textit{Compositional Complexity Index (CCI)}.

%\item \textbf{Open source.} We release %the codebase of the end-to-end %\textsc{DrBencher} pipeline and the human-%verified benchmark data at %\url{https://github.com/IBM/DrBencher}.

\end{itemize}

%The remainder of this paper covers related work (\S\ref{sec:related}), techniques and algorithms (\S\ref{sec:method}), experimental setup (\S\ref{sec:setup}), results (\S\ref{sec:results}), and conclusion (\S\ref{sec:conclusion}).

\section{Related Work}
\label{sec:related}

\textbf{Multi-hop question answering.}
HotpotQA~\citep{yang2018hotpotqa}, MuSiQue~\citep{trivedi2022musique},
and 2WikiMultihopQA~\citep{ho2020constructing} test multi-hop
retrieval over Wikipedia, while Bamboogle~\citep{press2023bamboogle}
and StrategyQA~\citep{geva2021strategyqa} target compositionality that
resists single-search shortcuts.  DROP~\citep{dua2019drop} tests
numerical reasoning over a given paragraph, requiring no browsing or
entity identification. A commonality of these datasets is that
questions are constructed \emph{from} pre-existing source
passages---each question is authored to be answerable from a given
Wikipedia paragraph or document. Our benchmark unifies these threads
while dispensing with such provenance text: agents
must identify an unnamed entity \emph{and} apply quantitative
reasoning to its properties through multi-hop retrieval, from questions
synthesized answer-first over SPARQL-selected entities and their KG
chains rather than derived from a seed passage.

\textbf{Knowledge-graph--based question generation.}
LC-QuAD~\citep{trivedi2017lcquad},
LC-QuAD~2.0~\citep{dubey2019lcquad2}, and
KQA~Pro~\citep{cao2022kqapro} generate questions by sampling SPARQL
templates over Wikidata or DBpedia. These benchmarks evaluate
structured-query translation rather than entity identification from
indirect clues or quantitative reasoning beyond the KG. Our pipeline
uses Wikidata differently: KG chains supply both entity-identification
clues and numerical properties, while domain-specific templates
perform the computation. To guard against KG noise, all clue facts
extracted from chains are verified against the corresponding Wikipedia
articles before question composition.

\textbf{Benchmarks for agentic and deep-research systems.}
GAIA~\citep{mialon2024gaia}, BrowseComp~\citep{wei2025browsecomp}, and
FRAMES~\citep{krishna2025frames} evaluate tool-augmented agents but
focus exclusively on retrieval---whether through web browsing or
retrieval-augmented lookup---without requiring computation. Our
benchmark demands both: browsing to resolve the unnamed entity and its
properties, and code execution to derive the numerical answer,
ensuring that neither skill alone suffices.

\textbf{Synthetic and dynamic benchmark generation.}
DyVal~\citep{zhu2024dyval} and LiveBench~\citep{white2024livebench}
generate fresh evaluation instances to resist
contamination. AutoBencher~\citep{li2025autobencher} optimizes for
salience, difficulty, and novelty, while Zero-shot
Benchmarking~\citep{pombal2025zsb} uses LLMs to both generate
synthetic test data and judge responses, requiring only two prompts
per task. These approaches rely on model-based scoring; our pipeline
instead produces programmatically verifiable gold answers.  Unlike
single-call benchmarks, our questions couple multi-hop entity
identification and property retrieval with quantitative reasoning,
and require tool-augmented agents to solve them.

\textbf{Deep-research agents and data construction.}
A rapidly growing line of work builds deep-research agents together
with pipelines that construct their training or evaluation data.
WebShaper~\citep{tao2025webshaper} formalizes information-seeking to
agentically synthesize data;
WebWatcher~\citep{geng2025webwatcher} extends deep research to the
vision--language setting; WebAggregator~\citep{wang2025webaggregator}
and Cognitive Kernel-Pro~\citep{fang2025cognitivekernelpro} curate
agentic data to train deep-research foundation models; and systems
such as MiroThinker~\citep{miromind2025mirothinker} and Tongyi
DeepResearch~\citep{tongyi2025deepresearch} scale open-web research
agents. On the evaluation side, Humanity's Last
Exam~\citep{phan2025hle} assembles expert-authored, closed-ended
questions at the frontier of human knowledge. \textsc{DrBencher} is
complementary to these efforts: rather than mining or aggregating real
web documents as provenance---or curating questions by hand---it
generates provenance-free questions with programmatically verifiable
gold answers and explicit complexity calibration, exercising the
joint entity-identification, property-retrieval, and computation
skills that these agent systems are designed to perform.

\textbf{Mathematical and scientific reasoning.}
GSM8K~\citep{cobbe2021gsm8k}, MATH~\citep{hendrycks2021math}, and
GPQA~\citep{rein2024gpqa} test mathematical and scientific reasoning
in isolation: each problem fully specifies all inputs. Our questions
\emph{withhold} the entity identity, coupling retrieval with reasoning
in a way that these computation-only benchmarks do not.

\begin{table*}[t]
\centering
\small
\renewcommand{\arraystretch}{0.85}
\begin{tabular}{@{}lcccccc@{}}
\toprule
\textbf{Benchmark}
  & \textbf{Multihop}
  & \textbf{HybridQ}
  & \textbf{Prog.\ gold}
  & \textbf{Syn}
  & \textbf{KG}
  & \textbf{Comp.\ calib.} \\
\midrule
HotpotQA~\citep{yang2018hotpotqa}      & \cmark & \xmark & \xmark & \xmark & \xmark & \xmark \\
MATH~\citep{hendrycks2021math}         & \xmark & \xmark & \cmark & \xmark & \xmark & \xmark \\
MuSiQue~\citep{trivedi2022musique}     & \cmark & \xmark & \xmark & \xmark & \xmark & \xmark \\
LC-QuAD~\citep{trivedi2017lcquad}      & \cmark & \xmark & \xmark & \xmark & \cmark & \xmark \\
GAIA~\citep{mialon2024gaia}            & \cmark & \xmark & \xmark & \xmark & \xmark & \xmark \\
AutoBencher~\citep{li2025autobencher}  & \xmark & \xmark & \xmark & \cmark & \xmark & \xmark \\
BrowseComp~\citep{wei2025browsecomp}   & \cmark & \xmark & \xmark & \xmark & \xmark & \xmark \\
%ZSB           & \xmark & \xmark & \xmark & \cmark & \xmark & \xmark \\
\midrule
\textbf{\textsc{DrBencher}}  & \cmark & \cmark & \cmark & \cmark & \cmark & \cmark \\
\bottomrule
\end{tabular}
\caption{Comparison of \textsc{DrBencher} with related benchmarks. HybridQ = each question jointly requires retrieval and computation, Prog.\ gold = programmatic gold answer, Syn = questions are synthetically generated, KG = clue facts verified against knowledge graph and Wikipedia, Comp.\ calib.\ = complexity calibration. \cmark\ = supported, \xmark\ = not supported.}
\label{tab:comparison}
\vspace{-2pt}
\end{table*}

% DrBencher method section -- \input from acl_latex.tex
% Unified algorithm: Phases 0, 1, 1.5, 2 + V1/V2 + Phase 3 (Claude) + diversity filter.
\vspace{-2ex}
% ===================================================================
\section{\textsc{DrBencher}'s Pipeline}
\label{sec:method}
% ===================================================================

The \textsc{DrBencher} pipeline follows an \emph{answer-first} design:
ground-truth data is fetched from authoritative sources, the gold
answer is computed \emph{before} the question is composed, and every
generated question passes through programmatic validation, two-stage
difficulty verification, and a diversity filter before entering
the final benchmark.

% -------------------------------------------------------------------
% Unified Algorithm
% -------------------------------------------------------------------
\begin{algorithm*}[!t]
\small
\caption{\textsc{DrBencher} Unified Algorithm.}\label{alg:drbencher}
\DontPrintSemicolon
\SetKwInput{KwIn}{Input}
\SetKwInput{KwOut}{Output}
\SetKwInput{KwParam}{Param}

\KwIn{Domain $D \in \{\texttt{bio}, \texttt{fin}, \texttt{geo}, \texttt{hist}, \texttt{sec}\}$,
      topic set $\mathcal{T}$, template bank $B$, tool stack $\mathcal{S}_D$}
\KwParam{V1 samples ($k_1$) / threshold ($\tau_1$);\ V2 ($k_2$, $\tau_2$)}
\KwOut{Final benchmark $Q^{*}$}

$Q^{*} \leftarrow \emptyset$ \;
\ForEach{topic $T \in \mathcal{T}$}{
  % Phase 0: Seed entity discovery & gold answer
  $\mathcal{E} \leftarrow \textsc{FetchEntities}(D, T)$ \;
  \ForEach{entity $e \in \mathcal{E}$}{
    $e.\texttt{chains} \leftarrow \textsc{FetchChains}(e, \text{hops}{=}2)$;\quad
    $e.\texttt{data} \leftarrow \textsc{FetchDomainData}(D, e)$ \;
    $t \leftarrow \textsc{SelectTemplate}(B, e)$;\quad
    \lIf{$t = \texttt{nil}$}{\textbf{continue}}
    $\boldsymbol{a^{*}}$ (gold answer) $\leftarrow \textsc{Execute}(t, e.\texttt{data})$\;
    \BlankLine
    % Phase 1 + 1.5: Derive clues from KG chains, verify against Wikipedia
    $F \leftarrow \textsc{DeriveClues}(e.\texttt{chains})$;\quad
    $F' \leftarrow \textsc{GroundViaWikipedia}(F, e.\texttt{chains})$ \;
    \lIf{$F$ does not span $\ge$3 chains or $|F'| < 3$}{\textbf{continue}}
    \BlankLine
    % Phase 2: Question composition
    $\boldsymbol{q}$ (question) $\leftarrow \textsc{LLM-Compose}(F', t)$;\quad
    \lIf{$q = \texttt{nil}$}{\textbf{continue}}
    add $(q, a^{*}, t, e)$ to candidates\;
  }
  \BlankLine
  % Phase 3: Programmatic QA validation
  \ForEach{candidate $(q, a^{*}, t, e)$}{
    discard if $\textsc{Execute}(t, e.\texttt{data}) \neq a^{*}$ \tcp*{answer mismatch}
    discard if $e.\texttt{name}$ or $a^{*}$ appears in $q$ \tcp*{entity/value leak}
    discard if $q$ is ambiguous \tcp*{non-unique answer}
  }
  \BlankLine
  % V1/V2: Two-stage difficulty verification
  \ForEach{survivor $(q, a^{*}, t, e)$}{
    discard if acc of $k_1$ closed-book samples $\ge \tau_1$ \tcp*{V1: too easy}
    discard if acc of $k_2$ samples from $\textsc{Agent}(q, \mathcal{S}_D) \ge \tau_2$ \tcp*{V2: tool-solvable}
  }
  $Q^{*} \leftarrow Q^{*} \cup$ survivors\;
}
$Q^{*} \leftarrow \textsc{DiversityFilter}(Q^{*})$\;
\textbf{return} $Q^{*}$\;
\vspace{-4pt}
\end{algorithm*}

\vspace{-2ex}
\subsection{Algorithm Walkthrough}
\label{sec:techniques}

We describe Algorithm~\ref{alg:drbencher} (see also Figure~\ref{fig:pipeline}) in detail. Given a \emph{topic} from a domain (e.g., mountains from geophysical, CVEs from security), the pipeline proceeds as follows.

\textbf{Phase~0: Entity Discovery and Reasoning-Context Selection
  (lines 3--7).}  We query the Wikidata SPARQL endpoint for entities
of the target type that possess quantitative properties (population,
area, elevation, coordinates, etc.). For each entity, we fetch
multi-hop knowledge-graph chains---two-hop paths radiating from the
entity through diverse predicates---and retrieve its numerical
properties.  A reasoning template is then selected from one of several
template families: currently \emph{quantitative modeling}
(e.g., exponential growth, population density, compound interest) and
\emph{scientific inference} (e.g., boiling point at altitude,
atmospheric pressure, great-circle distance), though new families can
be added without changing the pipeline.  The gold answer is computed
deterministically by executing the template's formula over KG-sourced
values.

\textbf{Phase~1: Clue Fact Extraction (lines 8--9).}  An LLM extracts
one descriptive fact per Wikidata chain about the seed entity,
selecting specific, low-frequency predicates (e.g., ``hosted the 1964
Olympics,'' ``founded by Romulus'') over generic ones. Each fact is
grounded: we verify that the claimed triple exists in the chain data
and that the fact is supported by the corresponding Wikipedia
article. Facts that fail grounding are rejected. At least three
grounded facts spanning three distinct chains are required to proceed.

\textbf{Phase~2: Question Composition (lines 10--11).}  A second LLM
call composes a natural-language question that (i)~describes the
unnamed entity using the grounded clue facts from Phase~1, and
(ii)~poses the quantitative reasoning task from Phase~0. To solve the
question, an agent must: (1)~identify the entity from the clues,
(2)~retrieve its quantitative properties, and (3)~execute the
computation.

\textbf{Phase~3: Programmatic QA Validation (lines 12--15).}  Each
candidate question is subjected to three programmatic checks: the gold
answer is reproducible by re-executing the computation code, and the
entity name does not leak into the question text, and the question
should not be ambiguous. Questions that fail any check are discarded.

\textbf{Verification Stages V1/V2 (lines 16--18).}  Surviving
questions are subjected to two verification rounds. In V1, the
generating model is prompted $k_1$ times in closed-book mode (no
tools, no context); questions answered correctly above
threshold~$\tau_1$ are discarded as too easy. In V2, the model is
given access to domain tools, a browser, and a Python interpreter
across $k_2$ independent trials; questions solved above~$\tau_2$ are
discarded.

\textbf{Diversity Filter (line 20).}
A graph-based filter removes near-duplicate questions from the surviving pool, selecting an approximately maximum independent set over cosine-dissimilarity edges (Section~\ref{sec:diversity}).

\vspace{-1.2ex}
% -------------------------------------------------------------------
% Compositional Complexity Index
% -------------------------------------------------------------------
\subsection{Compositional Complexity Index}
\label{sec:cci}

\begin{figure*}[t]
\centering
\resizebox{\textwidth}{!}{%
\small
\begin{tikzpicture}[
  >=Stealth,
  qbox/.style={text width=9cm, font=\normalsize\itshape, align=left,
               draw, rounded corners=3pt, inner sep=3pt, fill=white},
  kbox/.style={draw, rounded corners=2pt, minimum height=1.4em,
               font=\normalsize, align=center},
  dbox/.style={draw, fill=black!8, rounded corners=2pt,
               minimum height=1.4em, font=\normalsize, align=center},
  elabel/.style={font=\normalsize\bfseries, anchor=west},
  cci/.style={font=\normalsize\bfseries, draw, rounded corners=2pt,
              fill=black!12, inner sep=2pt},
]
\useasboundingbox (0, 0.1) rectangle (20, 7.8);

%% --- Left panel: E=1, P=1 (CCI=2) ---
\node[elabel] at (0.3,7.4) {(a)~$E{=}1,\; P{=}1$};
\node[qbox] (qa) at (4.8,6.0)
  {``A pendulum clock calibrated at sea level is taken to the summit of
  a mountain that hosted the 1936 Winter Olympics and is the highest
  peak in Germany. How many seconds does it lose per day?''};

\node[elabel] at (0.3,4.4) {Entity:};
\node[kbox] (a_e1) at (3.0,3.9)
  {Browse clues $\to$ \textbf{Zugspitze}};

\node[elabel] at (0.3,3.0) {Property:};
\node[kbox] (a_p1) at (3.1,2.5)
  {P2044 (elevation) $\to$ $h = 2{,}962$\,m};

\node[cci] at (6.8,3.0) {CCI $= 1 + 1 = 2$};

\node[elabel] at (0.3,1.6) {Computation:};
\node[dbox] (a_d1) at (1.8,0.8) {$g_h{=}g_0(\frac{R}{R+h})^2$\\[1pt]\textnormal{$\to$ 9.801}};
\node[dbox] (a_d2) at (4.8,0.8) {$\sqrt{g_0/g_h}$\\[1pt]\textnormal{$\to$ 1.00047}};
\node[dbox] (a_d3) at (7.8,0.8) {$86400\!\times\!(\cdot{-}1)$\\[1pt]\textnormal{$\to$ \textbf{40.28\,s}}};
\draw[->, thick] (a_d1.east) -- (a_d2.west);
\draw[->, thick] (a_d2.east) -- (a_d3.west);

%% --- Vertical divider ---
\draw[gray, line width=0.4pt] (9.7,0.1) -- (9.7,7.8);

%% --- Right panel: E=1, P=2 (CCI=3) ---
\node[elabel] at (10.3,7.4) {(b)~$E{=}1,\; P{=}2$};
\node[qbox] (qb) at (14.8,6.0)
  {``This place's official language is Spanish, it is the capital of
  B\'{i}o B\'{i}o Province, and its province borders Loncopu\'{e}
  Department. Compute its population density in people/km$^2$.''};

\node[elabel] at (10.3,4.4) {Entity:};
\node[kbox] (b_e1) at (13.2,3.9)
  {Browse clues $\to$ \textbf{Los \'{A}ngeles, Chile}};

\node[elabel] at (10.3,3.0) {Properties:};
\node[kbox] (b_p1) at (12.5,2.3)
  {P1082 (population)\\[1pt]\textnormal{$\to$ 143{,}023}};
\node[kbox] (b_p2) at (15.5,2.3)
  {P2046 (area)\\[1pt]\textnormal{$\to$ 29.99\,km$^2$}};

\node[cci] at (17.0,3.0) {CCI $= 1 + 2 = 3$};

\node[elabel] at (10.3,1.4) {Computation:};
\node[dbox] (b_d1) at (14.5,0.7) {$143{,}023 \div 29.99$\\[1pt]\textnormal{$\to$ \textbf{4{,}769.02}}};

\end{tikzpicture}
}% end resizebox
\caption{\small CCI illustrated for two questions.
\textbf{(a)}~$E{=}1, P{=}1$ (CCI\,$=$\,2): the model identifies
one entity (Zugspitze) and retrieves one property (elevation).
\textbf{(b)}~$E{=}1, P{=}2$ (CCI\,$=$\,3): the model identifies
one entity (Los~\'{A}ngeles, Chile) but must retrieve \emph{two}
properties (population \emph{and} area) to apply the population density
template.}
\label{fig:cci_example}
\end{figure*}

We quantify the intrinsic complexity of each generated question with a
\emph{Compositional Complexity Index} $\mathrm{CCI}(q)=E+P$.

\textbf{Number of entities ($E$).}  The number of real-world entities
the agent must identify from multi-hop clue facts.  Single-entity
questions have $E{=}1$; comparative questions requiring two entities
have $E{=}2$.  For example, in the pendulum-drift question of
Figure~\ref{fig:cci_example} (a), $E{=}1$: the model must discover the
entity identity (Zugspitze) from the clues.

\textbf{Number of properties ($P$).}  The number of distinct Wikidata
property types the agent must retrieve per entity.  In
Figure~\ref{fig:cci_example} (a), $P{=}1$: the property elevation
($h{=}2{,}962$\,m) has to be retrieved. In
Figure~\ref{fig:cci_example} (b), two properties, population
\emph{and} area, have to be retrieved, i.e.~ $P{=}2$.

CCI is model-independent and correlates well with extrinsic difficulty, as
shown in Table~\ref{tab:cci}.

%Questions also vary in \emph{reasoning depth}---the number of
%sequentially dependent computational steps---from $d{=}1$ (a
%single formula) to $d{=}4$ (multi-step derivations composing
%equations). However, our evaluation shows that reasoning depth is a secondary
%difficulty factor: the CCI components $E$ and $P$ are the
%dominant predictors of model accuracy (Section~\ref{sec:results:cci}).

% -------------------------------------------------------------------
% Diversity Filter
% -------------------------------------------------------------------
\subsection{Diversity Filter}
\label{sec:diversity}

%The surviving questions may cluster around popular entities or similar
%templates.
We enforce question diversity via a graph-based filter that
approximates a \emph{maximum independent set} over a near-duplicate
graph. We embed each candidate question with
\texttt{all-MiniLM-L6-v2}~\citep{reimers2019sentencebert}~($d{=}384$) and $\ell_2$-normalize the
resulting vectors.  Let $\mathcal{C} = \{q_1, \dots, q_n\}$ be the
candidate questions and $\phi(q) \in \mathbb{R}^{384}$ the unit-norm
embedding of $q$.  We define the cosine dissimilarity $\delta(q_i, q_j) = 1
- \phi(q_i)^\top \phi(q_j)$ and construct an undirected graph $G =
(\mathcal{C}, E)$ where $(q_i, q_j) \in E$ iff $\delta(q_i, q_j) <
\tau_d$ (i.e., the pair is too similar).  %The filter then iteratively removes the most redundant node---the one connected to the largest number of near-duplicate neighbours (ties broken by the sum of cosine similarities to its neighbours)---and updates the adjacency, until no near-duplicate pairs remain among the survivors. The surviving nodes form a greedy minimum vertex cover---the largest subset inwhich every pair satisfies $\delta(q_i, q_j) \ge \tau_d$.  
The filter iteratively removes the most redundant question, the one connected to the largest number of near-duplicate neighbors (ties broken by total cosine dissimilarity to those neighbors) — and updates the adjacency graph until no near-duplicate pairs remain. This procedure approximates a minimum vertex cover over the near-duplicate graph; the surviving nodes form the complementary approximate maximum independent set, i.e., the largest subset in which every pair satisfies $\delta(q_i,q_j) \ge \tau_d$  ($\tau_d$ is set to  $0.3$).

%\begin{equation}
%  q^{*} = \arg\max_{q \in \mathcal{C}}
%           \deg_G(q),
%  \qquad
%  \mathcal{C} \leftarrow \mathcal{C} \setminus \{q^{*}\}.
%  \label{eq:graph_filter}
% \end{equation}

%\textbf{Post-hoc diversity audit.}
%We verify the filtered benchmark with Self-BLEU~\citep{zhu2018texygen} (lexical redundancy) and mean pairwise cosine dissimilarity (semantic spread).

%To avoid circular evaluation, the post-hoc audit uses three embedding models independent of the filtering model: BGE~\citep{xiao2023cpack}, Granite Embedding~\citep{awasthy2025granite}, and E5~\citep{wang2022e5}. Results are reported in Section~\ref{sec:results:distinctiveness}.

\section{Experimental Setup}
\label{sec:setup}

We instantiate the pipeline described in Section~\ref{sec:method} across five benchmark domains \emph{biochemistry, financial, geophysical, history, security}. All generation, verification, and judging use a single model: \texttt{gpt-oss-120b} \citep{openai2025gptoss}, a 120-billion-parameter Mixture-of-Experts tool-calling agentic language model served via vLLM~\citep{kwon2023vllm}. Model serving (Table~\ref{tab:model}), verification hyperparameters (Table~\ref{tab:verification}), KG parameters, and infrastructure details are provided in Appendices~\ref{sec:appendix:model}--\ref{sec:appendix:infra}.

\subsection{Topic Entities}
\label{sec:setup:topics}

Each domain defines a set of \emph{topic entities} that determine which seed entities enter the pipeline (see Appendix~\ref{sec:appendix:topics} for the full listing).
Regardless of domain, every entity is resolved to a Wikidata QID, which serves as the common key for KG-chain extraction and clue grounding.
The entry point varies by domain: geophysical categories issue SPARQL queries filtered by quantitative properties and minimum sitelink count (${\ge}20$); the other four domains resolve domain-specific identifiers---SEC EDGAR tickers, NVD CVE IDs, PubChem/UniProt accessions, or Wikidata temporal entities---to their corresponding QIDs.
Static seed lists curated from authoritative registries serve as fallbacks when dynamic discovery yields insufficient candidates.
All five domains converge to the same downstream pipeline: Wikidata QID $\to$ multi-hop KG chains $\to$ clue fact extraction $\to$ Wikipedia grounding; only the entry point to the QID differs. A new topic requires only a Wikidata type identifier or a seed list, with no changes to the pipeline. This allows the benchmark to scale both sample count and topical diversity as needed.

\subsection{Reasoning Templates}
\label{sec:setup:templates}

We instantiate reasoning templates across multiple template families (representative examples in Appendix~Table~\ref{tab:templates}). Template--entity compatibility is governed by which quantitative properties each entity type possesses: countries and cities are routed to quantitative-modeling templates (population, area, GDP), while mountains use elevation-based scientific-inference templates, planets use mass/radius templates, and cities with coordinate data are eligible for haversine distance calculations. Each entity may generate up to 3 questions from distinct templates, ensuring template diversity within a topic.

%We randomize parameters in templates to prevent answer memorization across runs, e.g., growth rates ($r$), time horizons ($t$), compounding frequencies ($n$), and reference cities for distance ($d$) calculations.

\subsection{Benchmark Domains}
\label{sec:setup:benchmarks}

All five domains share Wikidata and Wikipedia as base knowledge sources for entity discovery, KG-chain extraction, and fact grounding; domain-specific APIs supply the quantitative data needed for gold-answer computation (Appendix~\ref{sec:appendix:datasources}). Table~\ref{tab:bench-stats} summarizes the benchmark before human validations: composition after diversity filtering~(a), Compositional Complexity Index (CCI\,$=$\,$E{+}P$) and KG chain statistics~(b).
%and post-hoc diversity scores~(c).

\begin{table*}[t]
\small
\setlength{\tabcolsep}{1.5pt}
\begin{minipage}[t]{0.29\textwidth}
\centering
\textbf{(a)} Benchmark statistics\\[4pt]
\begin{tabular}{@{}lrrrr@{}}
\toprule
\textbf{Domain} & \textbf{QAs} & \textbf{Tmpl.} & \textbf{Top.} & \textbf{Art.} \\
\midrule
Biochem      &  53 & 21 & 20 &  491 \\
Financial    & 104 & 35 & 47 &  558 \\
Geophysical  &  76 & 33 & 26 &  405 \\
History      &  73 & 26 & 19 &  388 \\
Security     &  48 & 14 & 23 &  140 \\
\midrule
\textbf{Total} & 354 & 129 & 135 & 1982 \\
\bottomrule
\end{tabular}
\end{minipage}%
\hfill%
\begin{minipage}[t]{0.68\textwidth}
\raggedleft
\textbf{(b)} CCI and KG chain statistics\\[4pt]
\setlength{\tabcolsep}{1.5pt}
\begin{tabular}{@{}lrrrrrrrr@{}}
\toprule
\textbf{Domain} & \textbf{$E{=}1$} & \textbf{$E{=}2$} & \textbf{$\bar{P}$} & \textbf{CCI} & \textbf{$\overline{\text{CCI}}$} & \textbf{C/Q} & \textbf{H/C} & \textbf{H/Q} \\
\midrule
Biochem  & 31 & 22 & 1.5 & 2--4 & 2.9 & 10.9 & 1.0 & 10.8 \\
Financial  & 87 & 17 & 1.8 & 2--6 & 2.9 & 17.2 & 0.6 & 10.3 \\
Geophysical  & 61 & 15 & 1.2 & 2--6 & 2.5 & 13.0 & 0.6 &  7.9 \\
History & 51 & 22 & 1.7 & 2--10 & 3.1 & 4.3 & 0.8 &  3.5 \\
Security  & 27 & 21 & 2.0 & 2--5 & 3.4 & 13.2 & 0.4 &  5.4 \\
\midrule
\textbf{Total} & 257 & 97 & 1.7 & 2--10 & 2.9 & 11.7 & 0.7 &  7.6 \\
\bottomrule
\end{tabular}
%\hfill%
%\begin{minipage}[t]{0.22\textwidth}
%\centering
%\textbf{(c)} Post-hoc diversity\\[4pt]
%\begin{tabular}{@{}lrr@{}}
%\toprule
%  & \textbf{SB$\downarrow$} & \textbf{PD$\uparrow$} \\
%\midrule
%Biochem  & 0.36 & 0.72 \\
%Financial  & 0.51 & 0.66 \\
%Geophysical  & 0.35 & 0.72 \\
%History & 0.32 & 0.77 \\
%Security  & 0.40 & 0.53 \\
%\midrule
%\textbf{Tot.} & 0.41 & 0.87 \\
%\bottomrule
%\end{tabular}
\end{minipage}
\caption{\small Benchmark summary.
\textbf{(a)}~Tmpl.: unique templates; Top.: entity topics; Art.: grounding articles.
\textbf{(b)}~$E$: entities; $\bar{P}$: mean property lookups; CCI\,$=$\,$E{+}P$; C/Q: chains/question; H/C: hops/chain; H/Q: hops/question.
%\textbf{(c)}~SB: Self-BLEU (4-gram); PD: pairwise cosine dissimilarity.
}
\label{tab:bench-stats}
\label{tab:diversity}
\end{table*}
%\vspace{-2ex}

\section{Results}
\label{sec:results}
We evaluate \textsc{DrBencher} along three axes:
\emph{validity}---human evaluation of question-answer pair qualities
(Section~\ref{sec:results:human}); \emph{difficulty}---automatic
evaluation against 3 frontier and 3 open weight models
(Section~\ref{sec:results:frontier}); and \emph{diversity}---post-hoc
diversity score comparison to manually constructed benchmarks
\textit{BrowseComp+, MATH-500, GPQA}
(Section~\ref{sec:results:distinctiveness}).

\subsection{Human Evaluations}
\label{sec:results:human}

\paragraph{Annotator qualifications, annotation protocol and evaluation criteria}
We recruit seven expert annotators, each with over five years of professional experience in QA system evaluation and linguistic annotations such as semantic parsing and benchmark creation. Each of the five benchmark domains is independently assigned to two annotators, yielding two judgments per question. Annotators use a custom Streamlit-based annotation tool that, for each question, displays: (i)~the generated question text, (ii)~the gold answer, (iii)~the gold entity chain with entity identification and property lookup steps, (iv)~the computation code, and (v)~grounding Wikipedia articles. Annotators evaluate each QA pair on two axes:
\begin{enumerate}[nosep,leftmargin=*]
  \item Entity Identification Accuracy --- verify that the entity identified in the gold entity chain is correct given the clue facts.
  \item Question Clarity --- verify that the question is unambiguous and contains no erroneous descriptions.
\end{enumerate}
Each item receives a binary \emph{Correct}/\emph{Incorrect} verdict; \emph{Incorrect} items require a free-text comment describing the issue.
The annotation interface is shown in Figure~\ref{fig:annotation-ui} (Appendix~\ref{sec:appendix:annotation}).

\textbf{Accuracy and Error Analysis.}
Table~\ref{tab:human-acc} summarizes the results: of the 354 generated questions reviewed by annotators and adjudicated by the first author, 76\% are valid (84\% when stale database entries are excluded).
Error analysis reveals three primary error types: (i)~stale or incorrect KG data (DB, 30 errors, 35\% of all errors), concentrated in financial filings where XBRL values change quarterly; (ii)~ambiguous or non-unique entity descriptions (Amb, 24 errors, 28\%), where clue facts do not sufficiently distinguish the target entity; and (iii)~LLM hallucination (LLM, 19 errors, 22\%).
LLM hallucination manifests at two pipeline stages.  In Phase~1
(\emph{fact extraction}), the LLM reads a KG triple correctly but
overstates its semantics in natural language---e.g., Wikidata P703
(\emph{found in taxon}) records that a compound was \emph{detected} in
an organism, yet the LLM renders this as ``produced by,'' implying a
biosynthetic origin that does not hold (7 of the biochemistry errors).
In Phase~2 (\emph{question composition}), the LLM fabricates
relationships absent from any extracted fact---e.g., inventing ``twin
headlands'' between two unrelated capes, or attributing one entity's
properties to another in a comparative question.  Multi-hop property
flattening (MH, 4 errors) occurs when a property of an
\emph{intermediate} entity in a 2-hop chain is incorrectly attributed
to the \emph{target} entity.  For example, Wikidata records that
Nufenen Pass \emph{lies in} the Lepontine Alps, which in turn
\emph{share a boundary with} the Uri Alps.  The generated question
drops the intermediary range and instead states that Nufenen Pass
itself ``shares a boundary with the Uri Alps.''  

\textbf{Inter-annotator agreement.}
We compute inter-annotator agreement for all five domains with two
independent annotators per domain, pooled over 159 items; per-domain
scores are shown in Table~\ref{tab:iaa} (Appendix~\ref{sec:appendix:iaa}).
The overall raw agreement is 76.1\% with Krippendorff's $\alpha = 0.30$.
Following \cite{krippendorff2018content}, $\alpha < 0.667$ signals
insufficient reliability for unadjudicated labels, motivating
third-party adjudication: the first author reviewed all annotations---original
KG triples, Wikipedia grounding articles, and domain-API values, with a written
rationale recorded per item---and rendered the final verdict.
Categorizing all 38 disagreements (Table~\ref{tab:iaa-disagree},
Appendix~\ref{sec:appendix:iaa}) shows they concentrate in the
\emph{subjective} error categories---LLM hallucination and question
ambiguity---whereas objectively verifiable stale KG data accounts for
only 10.5\%. The low $\alpha$ thus reflects the intrinsic difficulty of
a low-base-rate error-detection task ($\sim$20--25\% errors, where a few
borderline judgment calls move $\alpha$ substantially) rather than
unreliable underlying data.

\begin{table}[t]
\centering
\small
\setlength{\tabcolsep}{3.5pt}
\begin{tabular}{@{}lrrrrrrrrr@{}}
\toprule
\textbf{Domain} & $n$ & \textbf{Err} & \textbf{DB} & \textbf{LLM} & \textbf{MH} & \textbf{Amb} & \textbf{Ent} & \textbf{Acc} & \textbf{Acc*} \\
\midrule
Geophysical   & 76 & 11 &  1 &  7 & 2 & 1 & 0 & 85.5 & 86.8 \\
History       & 73 & 11 &  4 &  2 & 1 & 3 & 1 & 84.9 & 90.4 \\
Biochemistry  & 53 & 16 &  2 &  7 & 1 & 3 & 3 & 69.8 & 73.6 \\
Security      & 48 & 13 &  1 &  3 & 0 & 7 & 2 & 72.9 & 75.0 \\
Financial     & 104 & 35 & 22 &  0 & 0 & 10 & 3 & 66.3 & 87.5 \\
\midrule
\textbf{Total} & \textbf{354} & \textbf{86} & \textbf{30} & \textbf{19} & \textbf{4} & \textbf{24} & \textbf{9} & \textbf{75.7} & \textbf{84.2} \\
\bottomrule
\end{tabular}
\caption{Human verification accuracy and error breakdown. $n$: questions after diversity filtering; Err: total errors; Acc: accuracy (\%); Acc*: accuracy excluding DB errors (\%).
Error categories---DB: incorrect/stale KG data; LLM: LLM hallucination in fact extraction or question composition; MH: multi-hop property flattening; Amb: ambiguous or non-unique entity description; Ent: entity misidentification.}
\label{tab:human-acc}
\end{table}

\subsection{Automatic Evaluations}
\label{sec:results:frontier}

We evaluate three proprietary (Claude
Opus~4.6~\citep{anthropic2026opus46},
Gemini~2.5~Flash~\citep{comanici2025gemini25},
GPT-5.2~\citep{openai2025gpt52}) and three open-weight
(Llama~4~Maverick~\citep{meta2025llama4},
Qwen3-30B-A3B~\citep{yang2025qwen3},
Mistral-Small-3.2-24B~\citep{mistral2025small32}) models on the 268
human-validated questions. All evaluations use a
three-sample protocol ($n{=}3$) with default API temperature and
domain-appropriate tolerance (2\% relative, exact match for
history). Table~\ref{tab:frontier-all} reports accuracy; the
evaluation prompt and inference configuration are in
Appendix~\ref{sec:appendix:eval}; pairwise McNemar's tests are in
Appendix~\ref{sec:appendix:significance}. The best frontier model
achieves only 20\% answer accuracy across the 5 domains, suggesting
that the benchmark is quite challenging.
The gap between entity
identification and answer accuracy---86\% vs.\ 20\% for the best
model---indicates that property retrieval and computation, not entity
identification, are the primary bottlenecks.

\textbf{Failure analysis.}
To localize where the pipeline breaks, we conduct a two-stage error
analysis on all six models over the 268 questions (majority vote, 3
runs; Table~\ref{tab:failure-analysis}, Appendix~\ref{sec:appendix:failure-analysis}).  Stage~1 groups each failure
by relative error magnitude and entity correctness; Stage~2 inspects
the model's raw reasoning trace to split ``close but wrong'' answers
(2--50\% relative error) into incorrect property values versus genuine
arithmetic mistakes.  \emph{Property retrieval is the dominant failure
mode across every model} (39--64\% of failures): the model identifies
the entity but recalls approximate or incorrect quantitative values
from parametric memory rather than the exact figures in the
authoritative source (SEC EDGAR XBRL, NVD/CVSS, PubChem, Wikidata).
\emph{True calculation errors are rare} (0--20\%)---for three of six
models, zero confirmed cases exist where correct values were combined
with an incorrect formula---and \emph{API/timeout/parsing errors are
essentially zero}.  Entity-identification failures scale inversely
with model capability (14\% of Opus~4.6's failures vs.\ 40\% of
Qwen3-30B's).  The gap between entity identification and answer
accuracy therefore reflects the difficulty of retrieving precise
quantitative properties, not arithmetic or infrastructure failures.

Table~\ref{tab:cci} reveals a monotonic decline in accuracy as CCI
increases.  A Jonckheere--Terpstra trend
test~\citep{jonckheere1954distribution,terpstra1952testing} confirms a
significant decreasing trend across CCI levels ($Z = 6.75$, $p <
0.001$, one-sided), corroborated by Spearman's $\rho = -0.22$ ($p <
0.001$, $N{=}268$).  The CCI trend largely explains the low answer
accuracy of financial and security domains.  Security averages
CCI$=$3.49, with 91\% of questions at CCI$\geq$3, followed by
financial CCI$=$2.84.  A residual domain-specific effect persists even
after controlling for CCI (Appendix~\ref{sec:appendix:cci-domain}): at
CCI$=$2, financial (6.0\%) and security (5.6\%) still trail
biochemistry (34.7\%) and history (37.5\%), likely reflecting the
obscurity of SEC EDGAR XBRL tags and NVD/CVSS vulnerability data in
LLM pretraining corpora.

\begin{table*}[t]
\centering
\small
\setlength{\tabcolsep}{3pt}
\begin{tabular}{@{}l cc cc cc cc cc cc@{}}
\toprule
& \multicolumn{2}{c}{\textbf{Biochem}}
& \multicolumn{2}{c}{\textbf{Financial}}
& \multicolumn{2}{c}{\textbf{Geophysical}}
& \multicolumn{2}{c}{\textbf{History}}
& \multicolumn{2}{c}{\textbf{Security}}
& \multicolumn{2}{c}{\textbf{All}} \\
\cmidrule(lr){2-3}\cmidrule(lr){4-5}\cmidrule(lr){6-7}\cmidrule(lr){8-9}\cmidrule(lr){10-11}\cmidrule(lr){12-13}
$n$ & \multicolumn{2}{c}{37} & \multicolumn{2}{c}{69} & \multicolumn{2}{c}{65} & \multicolumn{2}{c}{62} & \multicolumn{2}{c}{35} & \multicolumn{2}{c}{268} \\
\midrule
\textbf{Model}
  & Ent & Ans
  & Ent & Ans
  & Ent & Ans
  & Ent & Ans
  & Ent & Ans
  & Ent & Ans \\
\midrule
Claude Opus 4.6 & \textbf{80.2} & \textbf{35.1} & \textbf{94.2} & \textbf{10.6} & 77.4 & 20.0 & \textbf{82.8} & \textbf{30.6} & \textbf{98.1} & \textbf{4.8} & \textbf{86.1} & \textbf{20.1} \\
Gemini 2.5 Flash & 57.7 & 28.8 & 88.9 & 7.7 & 72.3 & \textbf{26.7} & 79.6 & 24.7 & 81.0 & 1.9 & 77.4 & 18.4 \\
GPT-5.2 & 67.6 & 14.4 & 93.7 & 3.9 & \textbf{79.5} & 10.3 & 77.4 & 22.0 & 87.6 & 0.0 & 82.1 & 10.6 \\
\midrule
Llama 4 Maverick & 64.0 & 27.9 & 82.1 & 6.3 & 69.7 & 21.5 & 81.7 & \textbf{30.6} & 74.3 & 1.0 & 75.5 & 17.9 \\
Qwen3-30B-A3B-Thinking & 55.0 & 14.4 & 80.7 & 2.9 & 39.0 & 5.6 & 68.3 & 24.2 & 70.5 & 1.9 & 62.8 & 10.0 \\
Mistral-Small-3.2-24B-Instruct & 59.5 & 10.8 & 81.2 & 2.4 & 59.5 & 2.1 & 71.5 & 18.3 & 81.0 & 0.0 & 70.6 & 6.8 \\
\midrule
\textbf{Avg} & 64.0 & 21.9 & 86.8 & 5.6 & 66.2 & 14.4 & 76.9 & 25.1 & 82.1 & 1.6 & 75.7 & 14.0 \\
\bottomrule
\end{tabular}
\caption{Entity identification (Ent) and answer accuracy (Ans) (\%) across five \textsc{DrBencher} domains. Top: proprietary models; bottom: open-weight. Each model accuracy is the average of 3 runs.
McNemar's test (All, $N{=}268$, majority vote over 3 runs): Opus, Gemini, and Llama form a statistical tie (pairwise p$>$0.05); Opus significantly outperforms GPT5.2, Qwen, and Mistral at p$<$0.001.}
\label{tab:frontier-all}
\end{table*}

\begin{table*}[t]
\centering
\begin{minipage}{\textwidth}
\small
\begin{minipage}[b]{0.56\textwidth}
\raggedright
\setlength{\tabcolsep}{2pt}
\begin{tabular}{@{}cccrrrrrrrr@{}}
\toprule
$E$ & $P$ & CCI & $n$ & Op. & Gem. & GPT & Llm. & Qwn. & Mst. & \textbf{Avg} \\
\midrule
1 & 1 & 2 & 137 & 28.2 & 23.8 & 15.3 & 24.1 & 12.2 & 9.0 & \textbf{18.8} \\
1 & 2 & 3 & 54 & 14.2 & 13.0 & 5.6 & 16.7 & 8.6 & 8.6 & \textbf{11.1} \\
$\geq$2 & $\geq$2 & $\geq$4 & 77 & 10.0 & 12.6 & 5.6 & 7.8 & 6.9 & 1.7 & \textbf{7.4} \\
\bottomrule
\end{tabular}
\captionof{table}{Accuracy (\%) by entities ($E$) and data lookups ($P$) ($N{=}268$). CCI\,$=$\,$E{+}P$. Each model accuracy is the average of 3 runs. A Jonckheere--Terpstra test confirms a significant decreasing trend in accuracy across CCI levels ($Z = 6.75$, $p < 0.001$, one-sided), with Spearman's $\rho = -0.22$ ($p < 0.001$, $N{=}268$).}
\label{tab:cci}
\end{minipage}%
\hfill
\begin{minipage}[b]{0.41\textwidth}
\raggedleft
\setlength{\tabcolsep}{2pt}
\begin{tabular}{@{}lrrrr@{}}
\toprule
 & \textbf{SB} & \multicolumn{3}{c}{\textbf{Dist.\,$\uparrow$}} \\
\cmidrule(lr){3-5}
\textbf{Benchmark} & $\downarrow$ & \textbf{BGE} & \textbf{Granite} & \textbf{E5} \\
\midrule
\textsc{DrBencher} & 0.384 & \textbf{.548} & \textbf{.295} & \textbf{.248} \\
BrowseComp+ & 0.248 & .498 & .275 & .229 \\
MATH-500 & 0.199 & .448 & .239 & .200 \\
GPQA & \textbf{0.168} & .446 & .261 & .206 \\
\bottomrule
\end{tabular}
\captionof{table}{Diversity comparison. SB: Self-BLEU. Dist.: mean pairwise cosine dissimilarity measured with three independent embeddings: BGE, Granite, E5.}
\label{tab:distinctiveness}
\end{minipage}
\end{minipage}
\end{table*}
\label{sec:results:cci}

\subsection{Benchmark Distinctiveness}
\label{sec:results:distinctiveness}

\looseness=-1
We evaluate \textsc{DrBencher}'s distinctiveness using Self-BLEU (lexical redundancy) and mean pairwise cosine dissimilarity (semantic spread). Because the diversity filter (Section~\ref{sec:diversity}) uses \texttt{all-MiniLM-L6-v2}, we measure post-hoc diversity with three \emph{independent} embedding models---BGE~\citep{xiao2023cpack}, Granite Embedding~\citep{awasthy2025granite}, and E5~\citep{wang2022e5}---to ensure the result is not an artifact of the filtering model. Table~\ref{tab:distinctiveness} compares \textsc{DrBencher} against BrowseComp-Plus~\citep{wei2025browsecomp} (multi-hop entity identification), MATH-500~\citep{hendrycks2021math} (mathematical reasoning), and GPQA~\citep{rein2024gpqa} (graduate-level science) using equal-sized samples ($N{=}198$, limited by the smallest GPQA diamond sample size).
Across all three embedding models, \textsc{DrBencher} achieves the highest mean pairwise cosine dissimilarity. Its higher Self-BLEU (0.384) reflects shared domain terminology without reducing semantic spread. Notably, \textsc{DrBencher} is synthetically generated, yet exceeds the diversity of all three manually curated benchmarks regardless of embedding model.

\subsection{Robustness to the Difficulty-Filter Model}
\label{sec:results:verifier}

A single model (\texttt{gpt-oss-120b}) drives both question generation
and the two-stage difficulty filter (Section~\ref{sec:setup}), raising a
generator--verifier confound.  To test this, we re-ran the filter
(V1/V2) with a different model
(Llama~4~Maverick~\citep{meta2025llama4}) while holding question
generation (Phases~0--3) fixed
(Table~\ref{tab:verifier-ablation}).
For the 89 questions retained by \emph{both} filters the
programmatically computed gold answers are identical (89/89), confirming
that answer correctness is a property of the generation pipeline, not of
the filter model; and 78.8\% of \texttt{gpt-oss-120b}'s retained
questions also survive the weaker Llama filter.  The
difficulty-filter model therefore behaves as a \emph{calibration
hyperparameter} that sets the hardness threshold, not an architectural
dependency that determines which questions or answers are valid.

\begin{table}[t]
\centering
\small
\setlength{\tabcolsep}{5pt}
\begin{tabular}{@{}lrrrrc@{}}
\toprule
\textbf{Domain} & \textbf{GPT} & \textbf{Llama} & \textbf{Shared} & \textbf{\%GPT$\subseteq$Llama} & \textbf{Gold} \\
\midrule
Biochem     & 21 & 41 & 16 & 76.2 & 16/16 \\
Financial   & 18 & 37 & 13 & 72.2 & 13/13 \\
Geophysical & 41 & 61 & 31 & 75.6 & 31/31 \\
History     & 21 & 34 & 18 & 85.7 & 18/18 \\
Security    & 12 & 20 & 11 & 91.7 & 11/11 \\
\midrule
\textbf{Total} & \textbf{113} & \textbf{193} & \textbf{89} & \textbf{78.8} & \textbf{89/89} \\
\bottomrule
\end{tabular}
\caption{Difficulty-filter ablation: \texttt{gpt-oss-120b} vs.\ Llama~4~Maverick as the V1/V2 verifier, with question generation held fixed. GPT/Llama: questions surviving each filter; Shared: retained by both; \%GPT$\subseteq$Llama: fraction of \texttt{gpt-oss-120b}'s set also in Llama's; Gold: matching gold answers among shared questions. Gold answers match exactly (89/89), confirming answer-level model independence.}
\label{tab:verifier-ablation}
\end{table}

\section{Conclusion and Future Work}
\label{sec:conclusion}

%We presented \textsc{DrBencher}, a synthetic benchmark generator requiring multi-hop entity identification, property retrieval, and quantitative reasoning across five domains. Human evaluation yields 76\% validity; frontier models average 19\% accuracy with monotonic difficulty scaling (CCI$=$2: 16\%, CCI$=$4: 8.7\%). The multi-skill coupling requirement exposes integrated-capability failures invisible to single-skill benchmarks.

%We presented \textsc{DrBencher}, a synthetic benchmark generator that produces questions requiring joint multi-hop entity identification, property retrieval and quantitative reasoning across five domains. Human evaluation yields 76\% validity (84\% excluding stale data); automatic evaluation shows the best performing frontier model (Claude Opus 4.6) achieves 19\% accuracy, with monotonic difficulty scaling as compositional complexity increases (CCI$=$2: 16\%, CCI$=$4: 8.7\%).
We presented \textsc{DrBencher}, a synthetic benchmark generator that
produces questions requiring joint multi-hop entity identification,
property retrieval and quantitative reasoning across five
domains. Human evaluation yields 76\% validity; automatic evaluation shows the best performing frontier model
(Claude Opus 4.6) achieves 20\% accuracy, with monotonic difficulty
scaling as Compositional Complexity Index (CCI) increases.  Diversity analysis
against manually constructed benchmarks (BrowseComp+, MATH-500, GPQA)
shows \textsc{DrBencher} achieves the highest semantic diversity.

\textbf{Future work.} \emph{(i)~Stronger automated validity
filters:} The residual invalidity is dominated by a few detectable
categories. Therefore, targeted post-hoc filters---re-executing the gold
computation code against current API values to catch stale answers,
programmatic Wikidata queries to enforce entity uniqueness, and an
independent LLM judge from a different model family to flag
hallucinated clue facts---could reduce the invalidity rate. \emph{(ii)~Silver-standard
training data:} Beyond the question and its verified answer, the pipeline also exposes the full solution trace. Each generated item is therefore a complete worked example, not just a
question--answer pair, so the pipeline can be run to produce large-scale post-training data.

%The multi-skill coupling requirement---correctly identifying entities \emph{and} retrieving their properties \emph{and} performing quantitative reasoning---exposes integrated-capability failures invisible to single-skill benchmarks. A model that excels at computation but lacks retrieval, or retrieves facts but cannot reason quantitatively, will fail despite high scores on isolated benchmarks. This provides a more holistic assessment of the integrated capabilities required by deep research agents.

\section*{Ethics Statement}
\label{sec:ethics}

All data used in this work is drawn from publicly available sources: Wikidata, Wikipedia, SEC EDGAR, NIST NVD, PubChem, and UniProt. No personally identifiable information is collected or generated by the benchmark pipeline.

Human evaluation (Section~\ref{sec:results:human}) was conducted by seven expert annotators with professional experience in QA evaluation and semantic annotation. Annotators participated voluntarily and were compensated at standard institutional rates.

The benchmark generator uses LLMs for question composition (Phase~2) and difficulty verification (V1/V2), as described in Sections~\ref{sec:method}--\ref{sec:setup}. All gold answers are computed deterministically by executing parameterized code against knowledge-graph values; no LLM output is used as a gold answer.

\section*{Limitations}
\label{sec:discussion:limitations}
The \textsc{DrBencher} pipeline uses gpt-oss-120b for question composition, QA validation, and difficulty filtering --- a practical choice rather than an architectural one. Any capable instruction-following model with tool access can fill these roles; the method itself is model-agnostic, and the reported numbers should be understood as calibrated against this particular model rather than as absolute hardness guarantees.

\textbf{Scope of reasoning skills.} \textsc{DrBencher} deliberately
targets three composable skills---indirect entity identification,
quantitative property retrieval, and domain-specific
computation---and its current template families (\emph{quantitative
modeling} and \emph{scientific inference}) cover a subset of the
reasoning that deep-research agents perform. It does not test
qualitative reasoning such as open-ended synthesis, argumentation, or
causal explanation, and our claims should be read as scoped to
browsing-plus-computation tasks rather than to deep research in
full generality. This scope is a design choice rather than a hard
limit: because gold answers are produced by executing parameterized
code over KG values, the pipeline admits only reasoning that can be
expressed as a deterministic computation. The template system is
nonetheless extensible---adding a new family (e.g., comparative
ranking, temporal reasoning, or causal inference expressible as
computation) requires only defining its computation function and
property mappings, leaving the rest of the pipeline unchanged.

\bibliography{custom,references}
\bibliographystyle{colm2026_conference}

\clearpage
\appendix

\section{Model Configuration and Verification Hyperparameters}
\label{sec:appendix:verification}
\label{sec:appendix:model}

Table~\ref{tab:model} lists the model serving configuration. Table~\ref{tab:verification} summarizes the verification-stage hyperparameters.

\begin{table}[h]
\centering
\small
\begin{tabular}{@{}lr@{}}
\toprule
\textbf{Parameter} & \textbf{Value} \\
\midrule
Architecture              & MoE, 120B parameters \\
Quantization              & MXFP4 \\
Precision                 & bfloat16 \\
Tensor parallel size      & 8 \\
GPU memory utilization    & 0.9 \\
Max sequence length       & 131{,}072 tokens \\
Chunked prefill           & Enabled (batch 2{,}048) \\
Prefix caching            & Enabled \\
Attention backend         & FlashAttention \\
CUDA graph capture        & Full + Piecewise \\
\bottomrule
\end{tabular}
\caption{Model serving configuration.}
\label{tab:model}
\end{table}

\begin{table}[h]
\centering
\small
\begin{tabular}{@{}lcc@{}}
\toprule
\textbf{Parameter} & \textbf{V1} & \textbf{V2} \\
\midrule
Samples per question ($n$)   & 10   & 10 \\
Sampling temperature          & 0.7  & 1.0 \\
Accuracy threshold ($\tau$)   & 0.5  & 0.5 \\
Max agentic iterations ($K$)  & ---  & 200 \\
Tool timeout                  & ---  & 60\,s \\
Answer tolerance ($\epsilon$) & \multicolumn{2}{c}{5\%} \\
\bottomrule
\end{tabular}
\caption{Verification hyperparameters. V1: closed-book; V2: agentic with tool calls. $\epsilon{=}5\%$ relative tolerance is used during generation-time difficulty filtering; a stricter 2\% tolerance is applied during evaluation scoring (Section~\ref{sec:results:frontier}).}
\label{tab:verification}
\end{table}

The V2 temperature ($T_2 = 1.0$) is set higher than V1 ($T_1 = 0.7$) to encourage diverse reasoning paths in the agentic loop, increasing the chance that at least one sample finds a successful strategy. The threshold $\tau = 0.5$ means a question is retained only if fewer than half the model's attempts produce the correct answer, ensuring substantial difficulty.

\section{Reasoning Templates}
\label{sec:appendix:templates}

We instantiate reasoning templates across multiple template families.
Table~\ref{tab:templates} shows representative examples;
template--entity compatibility is described in
Section~\ref{sec:setup:templates}.

\begin{table}[h]
\centering
\small
\setlength{\tabcolsep}{3pt}
\begin{tabular}{@{}lllll@{}}
\toprule
\textbf{Family} & \textbf{Template} & \textbf{Computation} & \textbf{Properties} & \textbf{Unit} \\
\midrule
Quantitative modeling & Exponential growth & $\lfloor P(1{+}r)^t \rceil$ & P1082 & people \\
Quantitative modeling & GDP per capita & $G / P$ & P2131, P1082 & USD \\
\addlinespace[2pt]
Scientific inference & Surface gravity & $GM/r^2$ & P2067, P2120 & m/s$^2$ \\
Scientific inference & Atmospheric pressure & $P_0 e^{-Mgh/RT}/1000$ & P2044 & kPa \\
\bottomrule
\end{tabular}
\caption{Representative reasoning templates (4 of 37)}
\label{tab:templates}
\end{table}

\section{Knowledge-Graph Parameters}
\label{sec:appendix:kg}

For each entity, we fetch up to 20 two-hop KG chains from Wikidata, with a $10\times$ over-fetch factor (up to 500 raw chains) to enable diversification. The SPARQL endpoint is queried with a 120-second timeout and exponential backoff (up to 5 retries with jitter). Up to 3 questions are generated per entity, each using a distinct reasoning template. Entities with fewer than 3 surviving chains after blacklist filtering are discarded.

\section{Infrastructure}
\label{sec:appendix:infra}

Each job requests a single node with 8 NVIDIA GPUs in exclusive-process mode and 500\,GB host memory. Per-topic jobs run in parallel as independent LSF batch jobs, with a final merge job that depends on all topic jobs completing. Wave-based scheduling limits concurrency to 2 concurrent topic jobs for Wikidata-sourced benchmarks, preventing SPARQL endpoint rate-limiting. Typical per-topic wall-clock time is 2--6 hours depending on the number of entities and the V2 agentic loop depth. The merge job completes in under 10 minutes.

\section{Topic Entities}
\label{sec:appendix:topics}

Each domain defines a set of topic categories that seed entity discovery.
Within each category, entities are resolved to Wikidata QIDs via domain-specific identifiers or SPARQL queries (Section~\ref{sec:setup:topics}).

\paragraph{Biochemistry (20 topics).}
neurotransmitters, analgesics, antineoplastics, antidepressants, antivirals, amino acids, sugars, nucleotides, antifungals, toxins, metabolites, steroids, kinases, proteases, transporters, cytokines, structural proteins, transcription factors, viruses, parasites.

\paragraph{Financial (20 topics).}
construction, chemicals, mining \& metals, hospitality \& travel, freight \& logistics, medical devices, restaurants \& dining, payments \& fintech, clean energy, enterprise software, e-commerce, regional banks, asset management, healthcare providers, agribusiness, cybersecurity, gaming \& casinos, specialty finance, apparel \& footwear, oil field services.

\paragraph{Geophysical (39 topics).}
mountains, volcanoes, cities, countries, islands, rivers, lakes, deserts, glaciers, buildings, towers, bridges, dams, waterfalls, peninsulas, caves, canals, cliffs, plateaus, passes, hills, ridges, valleys, craters, gorges, canyons, fjords, harbors, observatories, national parks, archaeological sites, oases, capes, lighthouses, monuments, stadiums, chimneys, minarets, wind turbines.

\paragraph{History (20 topics).}
empires, assassinations, sieges, religious events, coups, migrations, constitutions, independence movements, civil wars, genocides \& atrocities, economic crises, naval battles, peace accords, scientific institutions, technological milestones, famines, liberation leaders, cold war events, world fairs \& olympics, trade routes.

\paragraph{Security (20 topics).}
container platforms, CI/CD tools, messaging systems, web servers, identity providers, VPN solutions, email servers, firewalls, SIEM tools, CMS platforms, programming runtimes, SCADA/ICS, endpoint security, mobile platforms, DNS services, package managers, IoT platforms, data breaches, ransomware families, load balancers.

\section{Data Sources}
\label{sec:appendix:datasources}

All five domains share the following knowledge-graph and encyclopedic sources for entity discovery, multi-hop chain extraction, and clue grounding:

\begin{itemize}[leftmargin=*,nosep]
  \item \textbf{Wikidata SPARQL Endpoint:} \url{https://query.wikidata.org/sparql}
  \item \textbf{Wikidata REST API:} \url{https://www.wikidata.org/w/api.php}
  \item \textbf{Wikipedia Action API:} \url{https://en.wikipedia.org/w/api.php}
\end{itemize}

\noindent Domain-specific APIs supply the quantitative data for gold-answer computation:

\paragraph{Biochemistry.}
\begin{itemize}[leftmargin=*,nosep]
  \item PubChem PUG REST API: \url{https://pubchem.ncbi.nlm.nih.gov/rest/pug/}
  \item UniProt REST API: \url{https://rest.uniprot.org/uniprotkb/}
  \item RCSB PDB REST API: \url{https://data.rcsb.org/rest/v1/}
  \item ChEMBL REST API: \url{https://www.ebi.ac.uk/chembl/api/data/}
\end{itemize}

\paragraph{Financial.}
\begin{itemize}[leftmargin=*,nosep]
  \item SEC EDGAR XBRL API: \url{https://data.sec.gov/api/xbrl/companyfacts/}
  \item SEC Company Tickers: \url{https://www.sec.gov/files/company_tickers.json}
\end{itemize}

\paragraph{Geophysical.}
Entity quantitative properties (elevation, coordinates, mass, height) are retrieved exclusively via the shared Wikidata SPARQL endpoint listed above.

\paragraph{History.}
Temporal properties (inception, dissolution, point in time) are retrieved via the shared Wikidata SPARQL endpoint listed above.

\paragraph{Security.}
\begin{itemize}[leftmargin=*,nosep]
  \item NIST NVD API v2.0: \url{https://services.nvd.nist.gov/rest/json/cves/2.0}
  \item FIRST EPSS API: \url{https://api.first.org/data/v1/epss}
  \item CISA KEV Feed: \url{https://www.cisa.gov/sites/default/files/feeds/known_exploited_vulnerabilities.json}
\end{itemize}

\section{Human Annotation Interface}
\label{sec:appendix:annotation}

Figure~\ref{fig:annotation-ui} shows a representative page from the Streamlit-based annotation tool used for human verification (Section~\ref{sec:results:human}).
For each QA pair, the annotator sees the generated question, the gold answer, the gold entity chain (with entity identification and property lookup steps), the computation code that derives the answer, and grounding Wikipedia articles with semantic highlighting of question-relevant (blue) and answer-relevant (green) passages.
The annotator assigns a binary verdict (\emph{Correct} or \emph{Incorrect}) and may add a free-text comment for rejected items.

%\begin{figure*}[t]
\begin{figure*}
\centering
\footnotesize
% ---- Outer frame mimicking the app window ----
\begin{tcolorbox}[
  colback=white, colframe=gray!50, boxrule=0.5pt,
  arc=3pt, left=4pt, right=4pt, top=4pt, bottom=4pt,
  width=\textwidth,
  title={\small\bfseries Item 17 / 59 \quad \texttt{geophysical / mountains}},
  fonttitle=\small\bfseries,
  coltitle=black, colbacktitle=gray!10, toptitle=2pt, bottomtitle=2pt,
]

% ---- Question ----
\begin{tcolorbox}[
  title={\small\bfseries Question},
  colback=blue!3, colframe=blue!35, boxrule=0.4pt,
  arc=2pt, left=4pt, right=4pt, top=2pt, bottom=2pt,
  fonttitle=\small\bfseries, coltitle=black,
  colbacktitle=blue!10, toptitle=1pt, bottomtitle=1pt,
]
\small\itshape
A towering figure created in the Neoclassicism movement, whose internal structure is made from steel, offers a view of Lower Manhattan.
If a simple pendulum were suspended from the top of this structure, what would be its period of oscillation in seconds?
Round your answer to two decimal places.
\end{tcolorbox}

\vspace{4pt}

% ---- Gold Answer ----
\begin{tcolorbox}[
  title={\small\bfseries Gold Answer},
  colback=green!4, colframe=green!45!black, boxrule=0.4pt,
  arc=2pt, left=4pt, right=4pt, top=2pt, bottom=2pt,
  fonttitle=\small\bfseries, coltitle=black,
  colbacktitle=green!12, toptitle=1pt, bottomtitle=1pt,
]
\textbf{13.61}~seconds
\end{tcolorbox}

\vspace{4pt}

% ---- Two-column: Entity Chain + Computation ----
\begin{minipage}[t]{0.52\textwidth}
\begin{tcolorbox}[
  title={\small\bfseries Gold Entity Chain},
  colback=gray!3, colframe=gray!45, boxrule=0.4pt,
  arc=2pt, left=4pt, right=4pt, top=2pt, bottom=2pt,
  fonttitle=\small\bfseries, coltitle=black,
  colbacktitle=gray!15, toptitle=1pt, bottomtitle=1pt,
  height=3.8cm, valign=top,
]
\small
\textbf{Entity Identification:}\\
\quad Statue of Liberty (\texttt{Q9202})\\[3pt]
\textbf{Property Lookup:}\\
\quad height (P2048) = \textbf{46.0\,m} \;[Wikidata]\\[3pt]
\textbf{Computation:}\\
\quad $T = 2\pi\sqrt{h/g}$ = \textbf{13.61\,s}
\end{tcolorbox}
\end{minipage}%
\hfill
\begin{minipage}[t]{0.46\textwidth}
\begin{tcolorbox}[
  title={\small\bfseries Computation Code},
  colback=yellow!3, colframe=gray!45, boxrule=0.4pt,
  arc=2pt, left=4pt, right=4pt, top=2pt, bottom=2pt,
  fonttitle=\small\bfseries, coltitle=black,
  colbacktitle=yellow!15, toptitle=1pt, bottomtitle=1pt,
  height=3.8cm, valign=top,
]
\small\ttfamily
import math\\
h = 46.0~~\# height (m)\\
g = 9.81~~\# gravity\\
T = 2 * math.pi * math.sqrt(h/g)\\
print(round(T, 2))~~\# 13.61
\end{tcolorbox}
\end{minipage}

\vspace{4pt}

% ---- Grounding Evidence ----
\begin{tcolorbox}[
  title={\small\bfseries Grounding Evidence},
  colback=gray!3, colframe=gray!45, boxrule=0.4pt,
  arc=2pt, left=4pt, right=4pt, top=2pt, bottom=2pt,
  fonttitle=\small\bfseries, coltitle=black,
  colbacktitle=gray!15, toptitle=1pt, bottomtitle=1pt,
]
\small
\textbf{[C3\_F1]}\; \colorbox{blue!10}{The Statue of Liberty} was created in the \colorbox{blue!10}{Neoclassical} artistic movement.\\[2pt]
\textbf{[C8\_F1]}\; \colorbox{blue!10}{The Statue of Liberty} has a steel internal structure designed by Gustave Eiffel.\\[2pt]
\textbf{[C18\_F1]}\; From the pedestal of \colorbox{blue!10}{the Statue of Liberty}, visitors can see \colorbox{blue!10}{Lower Manhattan}.\\[2pt]
\textbf{[P2048]}\; Height = \colorbox{green!12}{46.0\,m} \;[Wikidata]
\end{tcolorbox}

\vspace{4pt}

% ---- Annotation verdict ----
\begin{tcolorbox}[
  title={\small\bfseries Annotation},
  colback=white, colframe=gray!40, boxrule=0.4pt,
  arc=2pt, left=4pt, right=4pt, top=3pt, bottom=3pt,
  fonttitle=\small\bfseries, coltitle=black,
  colbacktitle=gray!10, toptitle=1pt, bottomtitle=1pt,
]
\small
\textbf{Verdict:}\quad
{\large$\CIRCLE$}~Correct\qquad
{\large$\Circle$}~Incorrect
\qquad
\textbf{Comment:}~\fbox{\hspace{3.5cm}\strut}
\quad
\fbox{\textbf{Save}}\quad
\fbox{\textbf{Save \& Next}}
\end{tcolorbox}

\end{tcolorbox}
\caption{Representative page from the human annotation interface. The tool displays the question, gold answer, gold entity chain, computation code, and grounding evidence with semantic highlighting (\colorbox{blue!10}{blue}: question-relevant entities; \colorbox{green!12}{green}: answer-relevant values). Annotators assign a binary verdict and optional comment.}
\label{fig:annotation-ui}
\end{figure*}

\section{Inter-Annotator Agreement Details}
\label{sec:appendix:iaa}

Table~\ref{tab:iaa} reports inter-annotator agreement across all five domains with two independent annotators per domain.
Raw agreement is computed over all items with overlapping annotations; Krippendorff's $\alpha$ accounts for chance agreement and is computed using the coincidence matrix approach \citep{krippendorff2018content}.

\begin{table}[h]
\centering
\small
\setlength{\tabcolsep}{5pt}
\begin{tabular}{@{}lrcc@{}}
\toprule
\textbf{Domain} & $n$ & \textbf{Agr.} & $\alpha$ \\
\midrule
Geophysical   & 47 & 78.7 & 0.25 \\
History       & 30 & 86.7 & 0.43 \\
Biochemistry  & 32 & 68.8 & 0.10 \\
Security      & 11 & 81.8 & 0.56 \\
Financial     & 39 & 69.2 & 0.36 \\
\midrule
\textbf{Overall} & \textbf{159} & \textbf{76.1} & \textbf{0.30} \\
\bottomrule
\end{tabular}
\caption{Inter-annotator agreement across all five domains. $n$: questions with overlapping annotations from two annotators; Agr.: raw agreement (\%); $\alpha$: Krippendorff's alpha. Overall statistics are weighted by domain size.}
\label{tab:iaa}
\end{table}

Table~\ref{tab:iaa-disagree} categorizes all 38 inter-annotator
disagreements (of the 159 matched pairs), as discussed in
Section~\ref{sec:results:human}.

\begin{table}[h]
\centering
\small
\setlength{\tabcolsep}{5pt}
\begin{tabular}{@{}lrr@{}}
\toprule
\textbf{Disagreement source} & \textbf{Count} & \textbf{\%} \\
\midrule
LLM hallucination / factual error (only) & 18 & 47.4 \\
Question ambiguity / non-uniqueness (only) &  8 & 21.1 \\
Both ambiguity + hallucination            &  7 & 18.4 \\
Stale KG data                             &  4 & 10.5 \\
Unclassified                              &  1 &  2.6 \\
\midrule
\textbf{Total} & \textbf{38} & \textbf{100.0} \\
\bottomrule
\end{tabular}
\caption{Breakdown of all 38 inter-annotator disagreements (of 159 matched pairs). Disagreements concentrate in subjective categories (hallucination, ambiguity); objectively verifiable stale data accounts for only 10.5\%.}
\label{tab:iaa-disagree}
\end{table}

Counting mixed cases, LLM hallucination is implicated in 65.8\% of
disagreements and question ambiguity in 39.5\%, while stale KG data---the
most objectively verifiable category---accounts for only 10.5\%. These
are genuine judgment calls: whether a paraphrased KG triple overstates
its semantics into a hallucination, or whether the clue facts
\emph{uniquely} identify the target entity, depends on each annotator's
strictness threshold. The low $\alpha$ thus reflects the intrinsic
difficulty of an error-detection task with a low base rate
(${\sim}20$--$25$\% errors), where a handful of borderline judgments
moves $\alpha$ substantially. Biochemistry is weakest ($\alpha=0.10$):
7 of its 10 disagreements involve hallucination or ambiguity in
biomedical relationships (e.g., whether a compound is ``naturally
sourced'' from a plant)---the class of question where non-specialist
annotators are least confident and domain-expert validation would help
most.

\section{Automatic Evaluation Prompt and Inference Configuration}
\label{sec:appendix:eval}

All models receive the same system prompt and inference configuration via LiteLLM~\citep{litellm2024}.
Table~\ref{tab:eval-config} lists the key inference parameters.
The system prompt instructs each model to reason step by step, identify the unnamed entity, and return a structured response:

\begin{quote}
\small\ttfamily
You are solving a benchmark question. The question describes one or more unnamed entities using factual clues. Think step by step, then give your response in the exact format below (each on its own line):\\
ENTITY: <the entity or entities you identified, comma-separated>\\
ANSWER: <your numerical answer>\\
The ENTITY line should contain the name(s) of the real-world entity (e.g., compound, company, country, person) you identified from the clues.\\
The ANSWER should be a single number --- no units, no explanation.
\end{quote}

\noindent Answers are extracted by parsing the \texttt{ANSWER:} tag from the model response (falling back to the last non-empty line if absent).
Entity identification is extracted from the \texttt{ENTITY:} tag and matched against gold entity names using normalized substring matching.

\begin{table}[h]
\centering
\small
\begin{tabular}{@{}lr@{}}
\toprule
\textbf{Parameter} & \textbf{Value} \\
\midrule
Samples per question ($n$) & 3 \\
Temperature & model default (omitted) \\
Max response tokens & 8{,}192 \\
Answer tolerance & 2\% relative (exact for history) \\
Retry on failure & 5 attempts, exponential backoff \\
\bottomrule
\end{tabular}
\caption{Evaluation inference configuration.}
\label{tab:eval-config}
\end{table}

\section{Two-Stage Failure Analysis}
\label{sec:appendix:failure-analysis}

Table~\ref{tab:failure-analysis} reports the two-stage failure analysis
discussed in Section~\ref{sec:results:frontier}: for each of the six
models, the distribution of failures over the 268 questions (majority
vote, 3 runs) across property-retrieval, calculation,
entity-identification, and infrastructure error categories.

\begin{table}[h]
\centering
\small
\setlength{\tabcolsep}{3pt}
\begin{tabular}{@{}lrrrrrr@{}}
\toprule
& \textbf{Opus} & \textbf{Gem.} & \textbf{GPT} & \textbf{Llm.} & \textbf{Qwn.} & \textbf{Mst.} \\
& 4.6 & 2.5 & 5.2 & 4~Mav & 3 & Sm. \\
\midrule
Answer acc.\ (\%)   &  20 &  17 &  10 &  16 &   8 &   6 \\
Failures ($n$)      & 214 & 222 & 242 & 224 & 247 & 251 \\
\midrule
\multicolumn{7}{@{}l}{\emph{Failure breakdown (\% of each model's failures):}} \\
Property retrieval       & \textbf{64.0} & 45.5 & \textbf{54.5} & \textbf{51.8} & 38.5 & \textbf{51.8} \\
True calculation         & 15.0 &  0.5 &  0.0 & 19.6 &  2.4 &  0.0 \\
Entity identification    & 14.0 & 22.1 & 17.8 & 25.9 & \textbf{39.7} & 29.9 \\
Property/formula (indet.)&  7.0 & 32.0 & 27.7 &  2.7 & 17.0 & 17.9 \\
API / timeout / parsing  &  0.0 &  0.0 &  0.0 &  0.0 &  2.4 &  0.4 \\
\bottomrule
\end{tabular}
\caption{Two-stage failure analysis over the 268 questions (majority vote, 3 runs). Rows below the rule are percentages of each model's failures. Property/formula (indet.): entity correctly identified and answer wrong, but the reasoning trace was insufficient to distinguish a wrong property value from a wrong formula. Models: Claude Opus~4.6, Gemini~2.5~Flash, GPT-5.2, Llama~4~Maverick, Qwen3-30B-A3B, Mistral-Small-3.2-24B.}
\label{tab:failure-analysis}
\end{table}

\section{Statistical Significance Tests}
\label{sec:appendix:significance}

We conduct pairwise McNemar's tests to assess whether observed differences in answer accuracy between models are statistically significant.
McNemar's test is appropriate for paired binary outcomes (correct/incorrect predictions on the same question set) and tests the null hypothesis that the two models have equal error rates.
Table~\ref{tab:mcnemar-all} reports p-values for all pairwise comparisons across all 268 evaluated questions.

\begin{table}[h]
\centering
\small
\setlength{\tabcolsep}{3pt}
\begin{tabular}{@{}lccccc@{}}
\toprule
 & Opus & Gemini & GPT5.2 & Llama & Qwen \\
\midrule
Gemini & 0.243 \\
GPT5.2 & <.001$^{***}$ & <.001$^{***}$ \\
Llama & 0.155 & 0.868 & 0.004$^{**}$ \\
Qwen & <.001$^{***}$ & <.001$^{***}$ & 0.404 & <.001$^{***}$ \\
Mistral & <.001$^{***}$ & <.001$^{***}$ & 0.095 & <.001$^{***}$ & 0.522 \\
\bottomrule
\end{tabular}
\caption{McNemar's test p-values for answer accuracy (All $N{=}268$). Per-question correctness is determined by majority vote across 3 runs. $^{**}$p$<$0.01, $^{***}$p$<$0.001.}
\label{tab:mcnemar-all}
\end{table}

The results show that Claude Opus~4.6 (20.1\% accuracy) significantly
outperforms GPT-5.2 (10.6\%, p<0.001), Qwen3-30B (10.0\%, p<0.001),
Mistral-Small-3.2 (6.8\%, p<0.001).  The differences between Opus and
Gemini~Flash (18.4\%, p=0.243), and Llama~Maverick (17.9\%, p=0.155)
are not statistically significant.  Among open-weight models, Llama 4
Maverick achieves the highest accuracy (17.9\%) and significantly
outperforms Mistral-Small-3.2 (p<0.001).

\section{CCI Distribution and Domain-Controlled Accuracy}
\label{sec:appendix:cci-domain}

Table~\ref{tab:cci-domain} shows the CCI distribution across domains.
Security has the highest mean CCI (3.49), with 91\% of questions at
CCI$\geq$3, while geophysical has the lowest (2.35), with 82\% at
CCI$=$2.  Security has the largest share of CCI$\geq$4 questions
(57\%). To determine whether the lower accuracy of financial and security is
explained solely by their higher CCI, Table~\ref{tab:cci-controlled}
compares per-domain accuracy at matched CCI levels.

\begin{table}[!ht]
\centering
\small
\setlength{\tabcolsep}{3pt}
\setlength{\abovecaptionskip}{4pt}
\setlength{\belowcaptionskip}{0pt}
\begin{tabular}{@{}lrcrrrrr@{}}
\toprule
\textbf{Domain} & $n$ & Mean & Med. & CCI=2 & CCI=3 & CCI=4 & CCI=6 \\
\midrule
Geophysical  &  65 & 2.35 & 2 & 53 &  3 &  8 & 1 \\
Biochemistry &  37 & 2.81 & 2 & 20 &  4 & 13 & 0 \\
History      &  62 & 2.82 & 3 & 24 & 25 & 13 & 0 \\
Financial    &  69 & 2.84 & 2 & 37 & 10 & 20 & 2 \\
Security     &  35 & 3.49 & 4 &  3 & 12 & 20 & 0 \\
\bottomrule
\end{tabular}
\caption{CCI distribution by domain. Mean: mean CCI; Med.: median CCI.}
\label{tab:cci-domain}
\end{table}
\vspace{-2.5ex}
\begin{table}[!ht]
\centering
\small
\setlength{\tabcolsep}{3pt}
\setlength{\abovecaptionskip}{4pt}
\setlength{\belowcaptionskip}{0pt}
\begin{tabular}{@{}crrrrrr@{}}
\toprule
CCI & Biochem & Financial & Geophysical & History & Security & All \\
\midrule
2 & 34.7 (20) &  6.0 (37) & 13.9 (53) & 37.5 (24) &  5.6 (3)  & 18.8 \\
3 &  6.9 (4)  &  6.1 (10) & 31.5 (3)  & 15.6 (25) &  2.3 (12) & 11.1 \\
$\geq$4 &  6.8 (13) &  4.8 (22) & 11.1 (9)  & 20.5 (13) &  0.6 (20) &  7.4 \\
\bottomrule
\end{tabular}
\caption{CCI-controlled accuracy (\%) by domain (number of questions
in parentheses).  Each cell is the average across 6 models $\times$ 3
runs.}
\label{tab:cci-controlled}
\end{table}

% ============================================================
% BOTH SAMPLES IN ONE FIGURE — placed early so LaTeX schedules
% it for the top of the next page (before the remaining text)
% ============================================================
\begin{figure*}[!t]
\centering
\scalebox{0.82}{\begin{minipage}{1.22\textwidth}
\footnotesize

% ---- tcolorbox style shortcuts ----
\tcbset{
  sampleouter/.style={colback=white, colframe=gray!50, boxrule=0.4pt, arc=2pt, left=2pt, right=2pt, top=2pt, bottom=2pt, width=\textwidth, fonttitle=\footnotesize\bfseries, coltitle=black, colbacktitle=gray!10, toptitle=1pt, bottomtitle=1pt},
  sampleq/.style={colback=blue!3, colframe=blue!35, boxrule=0.3pt, arc=1pt, left=2pt, right=2pt, top=1pt, bottom=1pt, fonttitle=\footnotesize\bfseries, coltitle=black, colbacktitle=blue!10, toptitle=1pt, bottomtitle=1pt},
  samplea/.style={colback=green!4, colframe=green!45!black, boxrule=0.3pt, arc=1pt, left=2pt, right=2pt, top=1pt, bottom=1pt, fonttitle=\footnotesize\bfseries, coltitle=black, colbacktitle=green!12, toptitle=1pt, bottomtitle=1pt},
  samplechain/.style={colback=gray!3, colframe=gray!45, boxrule=0.3pt, arc=1pt, left=2pt, right=2pt, top=1pt, bottom=1pt, fonttitle=\footnotesize\bfseries, coltitle=black, colbacktitle=gray!15, toptitle=1pt, bottomtitle=1pt, valign=top},
  samplecode/.style={colback=yellow!3, colframe=gray!45, boxrule=0.3pt, arc=1pt, left=2pt, right=2pt, top=1pt, bottom=1pt, fonttitle=\footnotesize\bfseries, coltitle=black, colbacktitle=yellow!15, toptitle=1pt, bottomtitle=1pt, valign=top},
  sampleev/.style={colback=gray!3, colframe=gray!45, boxrule=0.3pt, arc=1pt, left=2pt, right=2pt, top=1pt, bottom=1pt, fonttitle=\footnotesize\bfseries, coltitle=black, colbacktitle=gray!15, toptitle=1pt, bottomtitle=1pt},
}

% ================================================================
% (a) FINANCIAL
% ================================================================
\begin{tcolorbox}[sampleouter, title={\footnotesize\bfseries Financial Domain \quad \texttt{financial / pharmaceuticals}}]

\begin{tcolorbox}[sampleq, title={\footnotesize\bfseries Question}]
\footnotesize\itshape
The firm is headquartered in Thousand Oaks, California (Pacific Time Zone) and produces erenumab, a monoclonal-antibody medication for migraine. It is listed on both Nasdaq and the Hong Kong Stock Exchange. Based on the company's FY~2023 filing, what was its operating expense ratio (operating expenses / total revenue) as a percentage?
\end{tcolorbox}

\vspace{1pt}

\begin{tcolorbox}[samplea, title={\footnotesize\bfseries Gold Answer}]
\textbf{45.75}\%
\end{tcolorbox}

\vspace{1pt}

\begin{minipage}[t]{0.52\textwidth}
\begin{tcolorbox}[samplechain, title={\footnotesize\bfseries Gold Entity Chain}, height=2.2cm]
\footnotesize
\textbf{Entity:} Amgen Inc.\ (\texttt{Q470517}, AMGN)\\[1pt]
\textbf{Lookup:} SEC EDGAR XBRL (FY\,2023): Rev.\,=\,\$25.98\,B, CoR\,=\,\$6.45\,B, OI\,=\,\$7.64\,B\\[1pt]
\textbf{Formula:} $\text{OpEx}/R \times 100$; OpEx $= R - \text{CoR} - \text{OI}$
\end{tcolorbox}
\end{minipage}%
\hfill
\begin{minipage}[t]{0.46\textwidth}
\begin{tcolorbox}[samplecode, title={\footnotesize\bfseries Computation Code}, height=2.2cm]
\footnotesize\ttfamily
R~= 25979000000.0\\
CoR = 6454000000.0\\
OI = 7639000000.0\\
opex = R - CoR - OI\\
print(round(opex/R*100, 2)) \# 45.75
\end{tcolorbox}
\end{minipage}

\vspace{1pt}

\begin{tcolorbox}[sampleev, title={\footnotesize\bfseries Grounding Evidence}]
\footnotesize
\textbf{[C1]}\; \colorbox{blue!10}{Amgen} $\!\xrightarrow{\text{hq}}$ \colorbox{blue!10}{Thousand Oaks} $\!\xrightarrow{\text{tz}}$ \colorbox{blue!10}{Pacific TZ}\quad
\textbf{[C2]}\; \colorbox{blue!10}{Amgen} $\!\xrightarrow{\text{product}}$ \colorbox{blue!10}{erenumab} $\!\xrightarrow{\text{role}}$ \colorbox{blue!10}{mAb} $\!\xrightarrow{\text{treats}}$ \colorbox{blue!10}{migraine}\\[1pt]
\textbf{[C3]}\; \colorbox{blue!10}{Amgen} $\!\xrightarrow{\text{exchange}}$ \colorbox{blue!10}{Nasdaq};\; \colorbox{blue!10}{Amgen} $\!\xrightarrow{\text{exchange}}$ \colorbox{blue!10}{HKSE}\quad
\textbf{[XBRL]}\; Rev.\,=\,\colorbox{green!12}{\$25.98\,B}; CoR\,=\,\colorbox{green!12}{\$6.45\,B}; OI\,=\,\colorbox{green!12}{\$7.64\,B}
\end{tcolorbox}

\end{tcolorbox}

\vspace{3pt}

% ================================================================
% (b) SECURITY
% ================================================================
\begin{tcolorbox}[sampleouter, title={\footnotesize\bfseries Security Domain \quad \texttt{security / software}}]

\begin{tcolorbox}[sampleq, title={\footnotesize\bfseries Question}]
\footnotesize\itshape
The platform powers the UK's Care Quality Commission, supports multilingual content, and is built on a PHP-based framework that runs on Unix-like OSes. Using NVD data, determine how many CRITICAL-severity CVEs are listed for this product, then calculate the percentage of the critical attack surface reduced if 16 of those CVEs have been patched. Express as a percent to two decimal places.
\end{tcolorbox}

\vspace{1pt}

\begin{tcolorbox}[samplea, title={\footnotesize\bfseries Gold Answer}]
\textbf{39.02}\%
\end{tcolorbox}

\vspace{1pt}

\begin{minipage}[t]{0.52\textwidth}
\begin{tcolorbox}[samplechain, title={\footnotesize\bfseries Gold Entity Chain}, height=2.0cm]
\footnotesize
\textbf{Entity:} Drupal (\texttt{Q170855})\\[1pt]
\textbf{Lookup:} NVD --- 41 CRITICAL CVEs (of 1{,}336 total)\\[1pt]
\textbf{Formula:} Reduction $= 16/41 \times 100$
\end{tcolorbox}
\end{minipage}%
\hfill
\begin{minipage}[t]{0.46\textwidth}
\begin{tcolorbox}[samplecode, title={\footnotesize\bfseries Computation Code}, height=2.0cm]
\footnotesize\ttfamily
critical = 41; patched = 16\\
reduction = patched/critical*100\\
print(round(reduction, 2)) \# 39.02
\end{tcolorbox}
\end{minipage}

\vspace{1pt}

\begin{tcolorbox}[sampleev, title={\footnotesize\bfseries Grounding Evidence}]
\footnotesize
\textbf{[C1]}\; \colorbox{blue!10}{Drupal} $\!\xrightarrow{\text{used by}}$ \colorbox{blue!10}{CQC} $\!\xrightarrow{\text{country}}$ \colorbox{blue!10}{UK}\quad
\textbf{[C2]}\; \colorbox{blue!10}{Drupal} $\!\xrightarrow{\text{platform}}$ \colorbox{blue!10}{Symfony} $\!\xrightarrow{\text{lang}}$ \colorbox{blue!10}{PHP}\quad
\textbf{[C3]}\; \colorbox{blue!10}{Drupal} $\!\xrightarrow{\text{OS}}$ \colorbox{blue!10}{Unix-like}\\[1pt]
\textbf{[C4]}\; \colorbox{blue!10}{Drupal} $\!\xrightarrow{\text{lang.\ of work}}$ \colorbox{blue!10}{multilingual}\quad
\textbf{[NVD]}\; CRITICAL = \colorbox{green!12}{41};\; patched = \colorbox{green!12}{16}
\end{tcolorbox}

\end{tcolorbox}

\end{minipage}}% end scalebox
\caption{\footnotesize Representative QA pairs from the Financial (top) and Security (bottom) domains.
%Each panel shows the question with embedded multi-hop KG clues (\colorbox{blue!10}{blue}), the gold answer, the entity chain with data-source lookup, the computation code, and grounding evidence with answer-relevant values (\colorbox{green!12}{green}).
}
\label{fig:domain-samples}
\label{fig:sample-financial}
\label{fig:sample-security}
\end{figure*}

Even at matched CCI, financial and security remain 3--7$\times$ harder
than other domains.  At CCI$=$2, financial (6.0\%) and security
(5.6\%) trail biochemistry (34.7\%) and history (37.5\%).  At CCI$=$3,
the gap persists: security (2.3\%) and financial (6.1\%) versus
geophysical (31.5\%) and history (15.6\%).  This residual difficulty is likely to reflect
the obscurity of the underlying data sources---SEC EDGAR XBRL tags and
NVD/CVSS vulnerability scores---which are specialized structured data
rarely encountered in LLM pretraining corpora, making accurate property
retrieval particularly challenging regardless of question complexity.

\section{Benchmark Domain Samples}
\label{sec:appendix:samples}

Figure~\ref{fig:domain-samples} presents representative QA pairs from the Financial and Security domains, rendered in the same format used by the human annotation interface (Figure~\ref{fig:annotation-ui}).
Each panel shows the generated question, the gold answer, the gold entity chain with data-source lookup, the computation code, and the grounding evidence linking entity-identification clues (\colorbox{blue!10}{blue}) and answer-relevant values (\colorbox{green!12}{green}) to their KG sources.

\end{document}